\documentclass[10pt,journal,compsoc]{IEEEtran}

\ifCLASSOPTIONcompsoc
  \usepackage[nocompress]{cite}
\else
  \usepackage{cite}
\fi

\ifCLASSINFOpdf
\else
\fi
\usepackage{array}
\usepackage{marvosym}
\usepackage{tabularx}
\usepackage{amsmath}
\usepackage{graphicx}
\usepackage{algorithmic}
\usepackage{threeparttable}
\usepackage{titletoc}
\usepackage[ruled,linesnumbered]{algorithm2e}
\usepackage{cite}
\usepackage{amsfonts}
\usepackage{color}
\usepackage{colortbl}

\usepackage{bm}
\usepackage{enumitem}

\usepackage{bbding}
\usepackage{multirow}
\usepackage{indentfirst}
\usepackage{hyperref}   % hyperlinks
\usepackage{color,xcolor}
\definecolor{mygray}{gray}{.9}
\definecolor{dark-red}{rgb}{0.6, 0.1, 0.1}
\definecolor{dark-green}{rgb}{0, 0.6, 0.25} 
\definecolor{citecolor}{rgb}{0,0.443,0.737} 
\definecolor{linkcolor}{rgb}{0.956,0.298,0.235} 
\hypersetup{
    colorlinks=true,
    linkcolor=linkcolor,
    filecolor=magenta,
    citecolor=dark-green,
    urlcolor=dark-green % urlcolor= black,  % cyan
}

\newcommand{\black}[1]{{\color{black}#1}}
\usepackage{booktabs}
\usepackage{caption}
\begin{document}
% \title{Large-Scale 3D Medical Image Pre-training}

\title{Large-Scale 3D Medical Image Pre-training with Geometric Context Priors}

% \textsuperscript{\Letter}
\author{Linshan Wu, Jiaxin Zhuang, Hao Chen,~\IEEEmembership{Senior Member,~IEEE}% <-this % stops a space
        
\IEEEcompsocitemizethanks{\IEEEcompsocthanksitem Linshan Wu, Jiaxin Zhuang, and Hao Chen are with the Department of Computer Science and Engineering, The Hong Kong University of Science and Technology, Hong Kong, China.
E-mail: \href{mailto:linshan.wu@connect.ust.hk}{\black{linshan.wu@connect.ust.hk}}, \href{mailto:jzhuangad@cse.ust.hk}{\black{jzhuangad@cse.ust.hk}}, \href{mailto:jhc@cse.ust.hk}{\black{jhc@cse.ust.hk}}.
}% <-this % stops a space
\thanks{This paper is an extension of our CVPR 2024 paper\cite{VoCo}.}
\thanks{Corresponding author: Hao Chen (\href{mailto:jhc@cse.ust.hk}{\black{jhc@cse.ust.hk}}).}
}

% \markboth{Journal of \LaTeX\ Class Files,~Vol.~14, No.~8, August~2015}%
% {Shell \MakeLowercase{\textit{et al.}}: Bare Advanced Demo of IEEEtran.cls for IEEE Computer Society Journals}

\IEEEtitleabstractindextext{%
\begin{abstract}

The scarcity of annotations poses a significant challenge in medical image analysis, which demands extensive efforts from radiologists, especially for high-dimension 3D medical images.
Large-scale pre-training has emerged as a promising label-efficient solution, owing to the utilization of large-scale data, large models, and advanced pre-training techniques. However, its development in medical images remains underexplored. The primary challenge lies in harnessing large-scale unlabeled data and learning high-level semantics without annotations. 
We observe that 3D medical images exhibit consistent geometric context, \emph{i.e.}, consistent geometric relations between different organs, which leads to a promising way for learning consistent representations.
Motivated by this, we introduce a simple-yet-effective \textbf{Vo}lume \textbf{Co}ntrast (\textbf{VoCo}) framework to leverage geometric context priors for self-supervision. 
Given an input volume, we extract base crops from different regions to construct positive and negative pairs for contrastive learning. Then we predict the contextual position of a random crop by contrasting its similarity to the base crops.
In this way, VoCo implicitly encodes the inherent geometric context into model representations, facilitating high-level semantic learning without annotations. 
To assess effectiveness, we \textbf{(1)} introduce PreCT-160K, the largest medical image pre-training dataset to date, which comprises 160K Computed Tomography (CT) volumes covering diverse anatomic structures; \textbf{(2)} investigate scaling laws and propose guidelines for tailoring different model sizes to various medical tasks; \textbf{(3)} build a comprehensive benchmark encompassing 48 medical tasks, including segmentation, classification, registration, and vision-language. Extensive experiments highlight the superiority of VoCo, showcasing promising transferability to unseen modalities and datasets. VoCo notably enhances performance on datasets with limited labeled cases and significantly expedites fine-tuning convergence. Codes, datasets, and models are available at \href{https://github.com/Luffy03/Large-Scale-Medical}
{https://github.com/Luffy03/Large-Scale-Medical}.

% we \textbf{(1)} introduce the existing largest medical image pre-training dataset PreCT-160K, which comprises 160K Computed Tomography (CT) volumes covering diverse anatomic structures;

% Large-scale pre-training has exhibited remarkable efficacy as a label-efficient solution, owing to the utilization of large-scale data, large models, and advanced pre-training techniques.

% we \textbf{(1)} curate a large-scale dataset PreCT-160K, which comprises 160K Computed Tomography (CT) volumes for pre-training;

% we \textbf{(1)} curate the existing largest medical image pre-training dataset PreCT-160K, which comprises 160K Computed Tomography (CT) volumes covering the anatomic structures;

% Extensive experiments highlight the superior performances of VoCo, showcasing promising transferability to unseen modalities and datasets. VoCo notably enhances performance on datasets with limited labeled cases and significantly expedites training convergence.

% To unleash the power of both labeled and unlabeled data, we further develop a practical omni-supervised pre-training framework, which integrates self- and semi-supervised learning to leverage labeled and unlabeled medical images.
\end{abstract}

}

\maketitle
\IEEEdisplaynontitleabstractindextext
\IEEEpeerreviewmaketitle

\ifCLASSOPTIONcompsoc
\IEEEraisesectionheading{\section{Introduction}\label{sec:introduction}}
\else
\section{Introduction}
\label{sec:introduction}
\fi

\begin{figure*}
	\centering
	\includegraphics[width=1\linewidth]{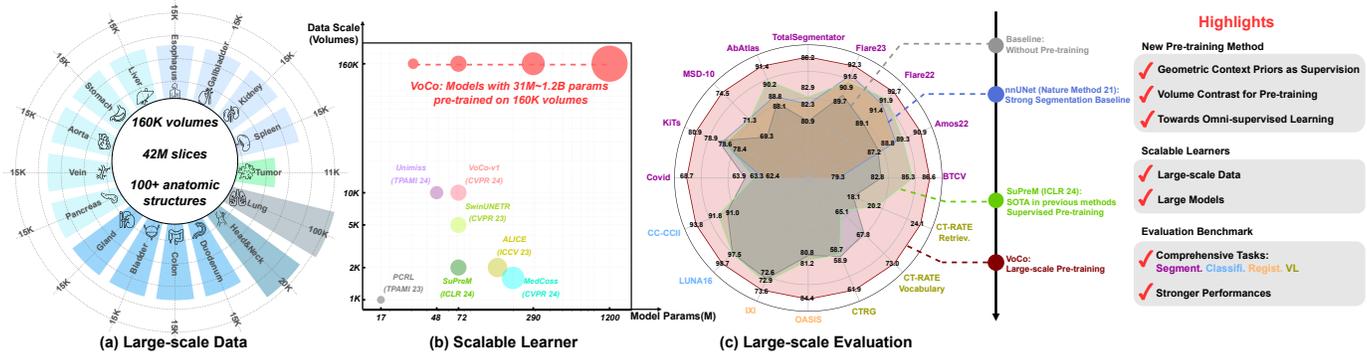}
	\caption{\textbf{Overview: (a)} We curate a large-scale 3D medical dataset PreCT-160K for pre-training. To the best of our knowledge, it is the existing largest pre-training dataset in this field, comprising 160K CT volumes (42M slices). \textbf{(b)} We investigate the scaling law in medical image pre-training, where VoCo stands out from previous methods in both data scale and model capacity. \textbf{(c)} We build a comprehensive benchmark for evaluation, which contains \textbf{48} downstream datasets across different tasks, \emph{i.e.}, segmentation, classification, registration, and vision-language (VL). Extensive experiments highlight the effectiveness of our proposed large-scale pre-training method.}
	\label{fig_teaser}
\end{figure*}

\IEEEPARstart{A}{I}-driven medical image analysis has witnessed emerging development in recent years~\cite{abdomenct1k,qiu2023rethinking,medley2021cycoseg,azad2024medical,xie2023learning,wu2022minimizing}, yet is heavily hampered by the high costs of the required expert annotations, especially for large-scale 3D medical images that with volumetric information~\cite{PCRLv2,UniMiSS+,SSL3D, annotating}. To address this dilemma, Self-Supervised Learning (SSL)~\cite{dino,ibot,simclr,mocov3,mim} for pre-training foundation models have demonstrated the potential to learn feature representations without the guidance from annotations, offering a promising solution in addressing the annotation bottleneck in 3D medical image analysis~\cite{PCRLv2,UniMiSS+,swin,Alice,continual}.

Recent advances~\cite{dino,dinov2,MAE,sam,internvl,depthanything} have highlighted the critical elements contributing to the success of vision foundation models, \emph{i.e.}, large-scale data, large models, and advanced pre-training techniques. However, \emph{\textbf{how well these solutions transfer to 3D medical image pre-training}} has not been thoroughly investigated. As shown in Fig.~\ref{fig_teaser}, \textbf{(1)} Data: previous methods~\cite{PCRLv1,PCRLv2,swin,SuPreM,unimiss,UniMiSS+,Alice,geo} are limited by the data scale (at most 10K volumes are used). Specifically, UniMiss~\cite{unimiss,UniMiSS+} innovatively proposed to boost chest CT pre-training by integrating 2D chest X-rays. However, the extendability to other anatomic regions remains under-explored. \textbf{(2)} Models: models trained in previous methods~\cite{PCRLv1,PCRLv2,swin,SuPreM,unimiss,UniMiSS+,Alice,geo} are still small-scale, with parameters only in the tens of millions. The scaling law of model capacity in medical image pre-training has not been well-explored. \textbf{(3)} Pre-training techniques: SuPreM~\cite{SuPreM} focused on supervised pre-training and annotated an abdomen segmentation dataset~\cite{atlas} for this purpose. Although showcasing state-of-the-art performance compared to previous methods, SuPreM~\cite{SuPreM} is still constrained by the scale of labeled data and fails to incorporate large-scale unlabeled data from diverse anatomical regions. In SSL, the majority of existing approaches~\cite{SSL3D,rubik,PCRLv1, PCRLv2,MAE3D,UniMiSS+,swin,continual} relied on low-level information reconstructions to learn augment-invariant representations, which typically employ data augmentation to the images and then reconstruct the raw information. However, the lack of high-level semantics in pre-training still impedes the performance of various downstream tasks. 

\begin{figure*}
	\centering
	\includegraphics[width=0.9\linewidth]{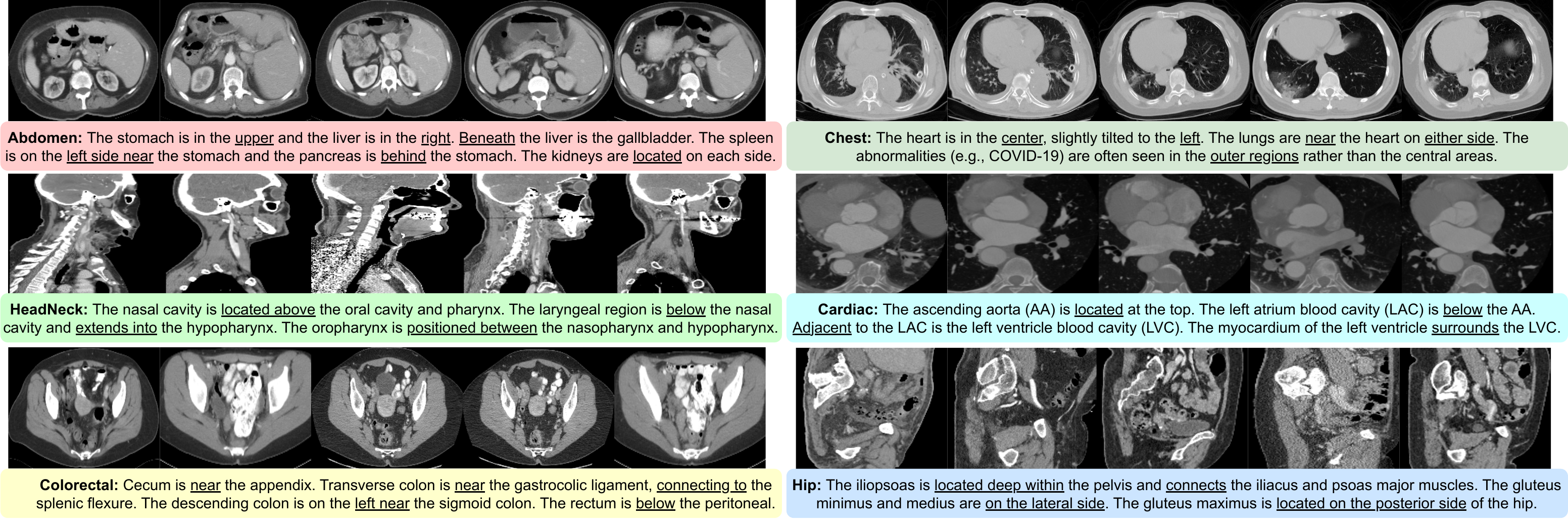}
    	\caption{\textbf{Motivation of VoCo}. In 3D medical images, the geometric relations between different organs are relatively consistent. We present some examples from PreCT-160K to illustrate these anatomical relationships across different regions. Motivated by this observation, we propose to leverage geometric context priors for learning consistent semantic representations and introduce a novel position prediction pretext task for pre-training.}
	\label{fig_prior}
\end{figure*}

The primary challenge is to incorporate high-level semantics for pre-training large-scale unlabeled data. We highlight that the geometric context priors of 3D medical images can be exploited. As illustrated in Fig.~\ref{fig_prior}, we observe that in 3D medical images, different organs (semantic regions) exhibit relatively consistent geometric relations with similar anatomic characteristics. Thus, the consistent geometric context between different organs offers a promising avenue for us to learn consistent semantic representations without the guidance of annotations in pre-training. 

In this paper, we propose a simple-yet-effective Volume Contrast (VoCo) framework, aiming to leverage the geometric context priors for contrastive learning. VoCo introduces a novel pretext task, \emph{i.e.}, contextual position predictions, aiming to encode the geometric relation of different organs into model representations. First, VoCo extracts a group of non-overlap base crops from different regions within an input volume. The base crops are employed to construct positive and negative pairs with a random crop for contrastive learning, \emph{i.e.}, base crops that overlap with the random crop are assigned as positive, otherwise negative. Then, we predict the contextual positions of a random crop by contrasting its similarity to the base crops. Intuitively, higher similarity indicates larger overlap areas, thus we can predict which region the random crop belongs to by calculating similarity. Specifically, we assign the overlap proportions between the random crop and base crops as position labels to supervise the position predictions. Through learning to predict contextual positions, VoCo implicitly encodes the inherent geometric contexts into the model representations without the guidance of annotations.

As shown in Fig.~\ref{fig_teaser}, existing works~\cite{PCRLv2,swin,SuPreM,UniMiSS+,Alice,continual}  are still constrained by the size of data, resulting in a large gap towards powerful medical vision foundation models. To this end, we curate a large-scale dataset PreCT-160K from publicly available sources, which currently stands as the largest and most comprehensive dataset for medical image pre-training. As shown in Fig.~\ref{fig_teaser}(a), PreCT-160K comprises over 160K CT volumes with an excess of 42M slices, encompassing the 3D anatomical map of human bodies. PreCT-160K also includes a substantial portion of labeled data, enabling us to combine self- and semi-supervised learning for omni-supervised pre-training. In this paper, we propose an omni-supervised pre-training framework to effectively unleash the power of labeled and unlabeled medical images.

We further explore the scaling law of model capacity and develop guidelines for tailoring different model sizes to diverse medical tasks. Specifically, we build a large-scale evaluation benchmark for medical image pre-training. In contrast to previous studies~\cite{PCRLv2,swin,SuPreM,UniMiSS+,Alice,continual} that were limited in evaluation data and tasks, our benchmark encompasses 48 downstream datasets spanning various tasks such as segmentation, classification, registration, and vision-language. Extensive experimental results on 48 downstream datasets demonstrate that our proposed VoCo significantly outperforms existing methods by a clear margin and achieves new state-of-the-art performances. 

The preliminary version of this study was presented in CVPR 2024~\cite{VoCo} and we named it VoCo-v1. In this paper, we made significant and substantial modifications, retaining the initial name as VoCo. The new contributions of this paper include but are not limited to:
\begin{itemize}

\item Compared to VoCo-v1~\cite{VoCo} that solely focused on intra-volume contrastive learning, we further introduce inter-volume contrastive learning with a momentum-based teacher-student module, enabling us to learn consistent representations between different volumes.

\item We investigate the combination of self- and semi-supervised learning for omni-supervised pre-training, effectively leveraging both labeled and unlabeled data.

\item We introduce the existing largest medical image pre-training dataset PreCT-160K and scale up the data scale from 10K~\cite{VoCo} to 160K. Our PreCT-160K is poised to foster future research in medical image pre-training.

\item We build the existing largest evaluation benchmark for medical image pre-training, encompassing diverse tasks across 48 downstream datasets. Our open-source implementation of various medical tasks will also benefit the following researchers in this field. 

\item We delve deeper into the scaling law and release pre-trained models with parameter sizes ranging from 31M to 1.2B. We also propose the guidelines for tailoring different model sizes to various medical tasks. 

\item We provide detailed and insightful analyses to underscore the core components of VoCo. These experiments further highlight the significance of large-scale pre-training, offering valuable insights that can inspire future research in the field of medical image pre-training.

\end{itemize}

\section{Related Work}

\subsection{Large-scale Vision Pre-training}

Vision pre-training opens up immense opportunities for harnessing large-scale vision data, playing a pivotal role in the development of large vision foundation models~\cite{dino,dinov2,sam,internvl,depthanything,depthv2}. The primary challenge lies in devising effective pre-training methodologies. Although supervised pre-training stands as a straightforward approach, it grapples with the challenge of the lack of manual annotations, demanding substantial engineering efforts for building labeled datasets. Deng et al.~\cite{imagenet} built the famous ImageNet and ImageNet pre-training has demonstrated its effectiveness in boosting downstream tasks. SAM~\cite{sam} introduced the SA-1B dataset with over 1 billion segmentation masks for supervised pre-training, thus achieving a strong segmentation foundation model. However, the high costs of annotations and the neglect of large-scale unlabeled data still hinder the further development of supervised pre-training. To this end, SSL proposed to learn robust features without the guidance of annotations~\cite{dino, dinov2, MAE, simclr, ibot}, which has garnered significant attention recently.

\textbf{Typical SSL methods}. SSL has showcased promising results across various vision tasks~\cite{dino, dinov2, MAE, swav, simclr, ibot}. DINO~\cite{dino,dinov2} proposed to integrate advanced SSL methods and learn robust features without annotations, which has become a prevalent choice for pre-trained backbones in contemporary research. State-of-the-art SSL methods can broadly be classified into generative~\cite{MAE,larsson2017colorization,pathak2016context,chen2020generative} and contrastive~\cite{simclr,swav,moco,byol,DCA,ibot,disco,agmm,simsiam} learning methods. \textbf{(1)} Generative learning methods are mainly based on reconstructing raw information from augmented images. For example, MAE~\cite{MAE} proposed to mask random patches of the input image and reconstruct the missing pixels. \textbf{(2)} Contrastive learning methods aim to learn consistent representations by contrasting positive and negative pairs of samples. 

\textbf{Transfer to medical image analysis}. Although the methods discussed above have achieved promising outcomes in natural images, directly applying these pre-trained models to medical images encounters challenges due to domain gaps~\cite{PCRLv2,UniMiSS+,unimiss,continual,geo,swin}. DINO~\cite{dino,dinov2} pre-trained a series of strong 2D vision Transformers~\cite{vit} and exhibited significant transferability in 2D medical images like X-ray and pathology images~\cite{chen2024towards,vorontsov2024foundation,vaidya2024demographic}. However, in the realm of challenging 3D medical tasks that necessitate volumetric information extraction, strong pre-trained 3D models are still under-explored~\cite{SuPreM,UniMiSS+,nnunet,PCRLv2}. 

Most state-of-the-art SSL methods~\cite{MAE,moco,mocov3,simclr,swav} often fall short in achieving competitive performances in 3D medical images, primarily caused by the ignorance of the unique characteristics of 3D medical images~\cite{UniMiSS+,PCRLv2,geo,Alice}. Specifically: \textbf{(1)} Contrastive learning~\cite{moco,mocov3} in natural images proposed to build positive and negative pairs of samples in a training batch, \emph{i.e}, the augmented view of input is assigned as positive, and other images are negative. However, for 3D medical images that share similar anatomical structures, it is difficult to build negative pairs in this way~\cite{swin,SSL3D,annotating,geo}. \textbf{(2)} Mask image modeling~\cite{MAE,ibot} proposed to mask and reconstruct missing pixels. However, for 3D medical images characterized by high dimensions, large sizes, and a significant background proportion, these methods often encounter issues as models tend to converge towards reconstructing irrelevant background regions~\cite{MAE3D,GLMAE,mim,swinmm,swin,Mis-fm}, diminishing the understanding of semantic regions (\emph{e.g.}, organs). Thus, the development of advanced SSL techniques for 3D medical images necessitates a meticulous consideration of the unique image information and the formulation of tailored strategies.

\subsection{Large-scale Medical Image Pre-training}

Medical image pre-training has been proven as an effective way to mitigate the scarcity of annotation in medical tasks~\cite{SuPreM,PCRLv2,UniMiSS+,Alice,he2024foundation,he2024meddr}. Early attempts~\cite{Modelgen,PCRLv1,TransVW} conducted pre-training on 2D X-ray images~\cite{wang2017chestx,chexpert}, demonstrating improvements on chest pathology identification and pneumothorax segmentation. In comparison, 3D medical images, \emph{e.g.}, CT and Magnetic Resonance Imaging (MRI), offer richer volumetric information for clinical diagnosis, which has received increasing attention in medical image analysis~\cite{SSL3D,annotating,nnunet,mim,SuPreM,meduniseg}. Nonetheless, the complexity inherent in 3D medical images introduces significant challenges to pre-training. Although recent works~\cite{PCRLv2,UniMiSS+,continual,SuPreM,Alice,swin,geo} have demonstrated the effectiveness of 3D medical image pre-training, significant challenges still persist, particularly in the realms of data scale, model capacity, and pre-training methods.

\subsubsection{Large-scale Data}

Compared with 2D X-ray, collecting 3D medical images like CTs is more difficult, stemming from factors such as slower imaging speeds, heightened radiation exposure, and increased costs~\cite{CT,withers2021x}. As shown in Fig.~\ref{fig_teaser}(b), most existing methods~\cite{PCRLv1,PCRLv2,swin,Alice,continual,SuPreM} leveraged limited scale of 3D data for pre-training. FreeTumor~\cite{freetumor} first investigated the data-scaling law in tumor segmentation with 11K CTs. Wang et al.~\cite{Mis-fm} built a dataset of 100K CTs for pre-training but it is not publicly available for research. To collect large-scale 3D data for pre-training, the necessity arises to \textbf{\emph{aggregate datasets from diverse sources}}, encompassing various hospitals across different regions and countries~\cite{atlas,embracing}. This will lead to diverse image characteristics and inconsistent imaging quality in the dataset, introducing new challenges to pre-training. 

Moreover, previous methods mainly collected data from specific body parts for pre-training, \emph{e.g.}, PCRL~\cite{PCRLv1,PCRLv2} and Unimiss~\cite{unimiss,UniMiSS+} on chest region, Alice~\cite{Alice} and SuPreM~\cite{SuPreM} on abdomen region, GVSL~\cite{geo} on cardiac region. However, given the distinct characteristics present in various anatomical regions, the transferability of models pre-trained on one region to another may be constrained~\cite{VoCo,mim,GLMAE}. In this paper, we build a large-scale dataset PreCT-160K that encompasses diverse anatomic structures. However, data sourced from different anatomical regions exhibit varying imaging parameters, \emph{i.e.}, different sizes, spacing, and intensities, posing new challenges for learning consistent representations in pre-training.

\subsubsection{Large Model}

Early works~\cite{Modelgen,TransVW,PCRLv1} in 3D medical image pre-training were constrained in model capacity, typically comprising only tens of millions of parameters. Recent advances~\cite{vit,dinov2,internvl,eva,MAE} have demonstrated the astonishing effectiveness of scaling law, where large models trained on large-scale data exhibit remarkable intelligence. In this paper, we collect a large-scale 3D medical image dataset, which comprises diverse image characteristics from various sources. The availability of such extensive data unlocks new opportunities for us to train large models.

Given the diversity of various medical tasks, it is imperative to evaluate large models on comprehensive benchmarks. Previous methods~\cite{Modelgen,TransVW,unimiss,PCRLv2,Alice,swin,SuPreM} primarily evaluated the pre-trained models on only a few downstream tasks, typically focusing on segmentation or classification tasks. STU-Net~\cite{stunet} was the first to evaluate large models yet is limited in segmentation tasks. In this paper, we delve deeper into the scaling law in various medical tasks, providing insights into tailoring diverse model sizes to accommodate varying medical tasks effectively.

\subsubsection{Advanced Pre-training Techniques}

\textbf{SSL for 3D medical images}. Existing methods~\cite{SSL3D, rubik, PCRLv2, MAE3D,continual,intra} are mostly based on information reconstructions to learn augment-invariant representations of 3D medical images, which first employ strong data augmentation to the images and then reconstruct the raw information. Rotate-and-reconstruct~\cite{SSL3D, swin, rubik, rubik2} proposed to randomly rotate the 3D volumetric images and learn to recover them, fostering the learning of rotation-invariant features. Recent methods~\cite{PCRLv1,PCRLv2,geo,dira,unimiss,Alice} delved into restoring low-level information across varied image perspectives. PCRL~\cite{PCRLv1, PCRLv2} cropped global and local patches then conducted multi-scale restorations. GVSL~\cite{geo} further explored the geometric similarities among multi-scans through affine augmentation and matching. Mask-reconstruct methods~\cite{MAE3D,GLMAE,swinmm,continual,mim} were derived from MAE~\cite{MAE}, aiming to learn representations by masking images and reconstructing the missing pixels. 
Although promising results have been demonstrated, the majority of these approaches often overlook the significance of integrating high-level semantics into model representations, thus impeding the further improvements in downstream tasks.

\textbf{High-level semantics in pre-training}. For medical images, high-level semantic information primarily stems from manual annotations, since it heavily relies on expert knowledge. Previous works~\cite{SuPreM,dodnet,clipdriven,stunet} proposed that supervised pre-training is more efficient and can achieve higher performances with less training time and labeled data~\cite{SuPreM}. However, the ongoing challenges persist in the scarcity of labeled data, impeding the transferability to various medical tasks, different anatomical structures, and extensive unseen datasets.
In this paper, we aim to integrate large-scale unlabeled data into pre-training. Thus, we propose to leverage the inherent characteristics of medical images as high-level semantic priors for self-supervision.

\textbf{Omni-supervised Learning}. Although self-supervised learning enables us to involve large-scale unlabeled data in pre-training~\cite{mim,VoCo,UniMiSS+}, it still overlooks the utilization of readily available labeled data. Omni-supervised learning~\cite{shu2023omni,DBFNet,tan2023positive,wu2024modeling} introduced the concept of leveraging diverse information for supervision. Specifically, semi-supervised learning~\cite{yang2022st++,CISC_R,yang2023revisiting,liu2023multi} demonstrates powerful efficacy in leveraging both labeled and unlabeled data. In this paper, we propose a simple-yet-effective omni-supervised pre-training framework, which combines self- and semi-supervised learning to unleash the power of both labeled and unlabeled medical images. 

\section{Method} 

\begin{figure}
	\centering
	\includegraphics[width=1\linewidth]{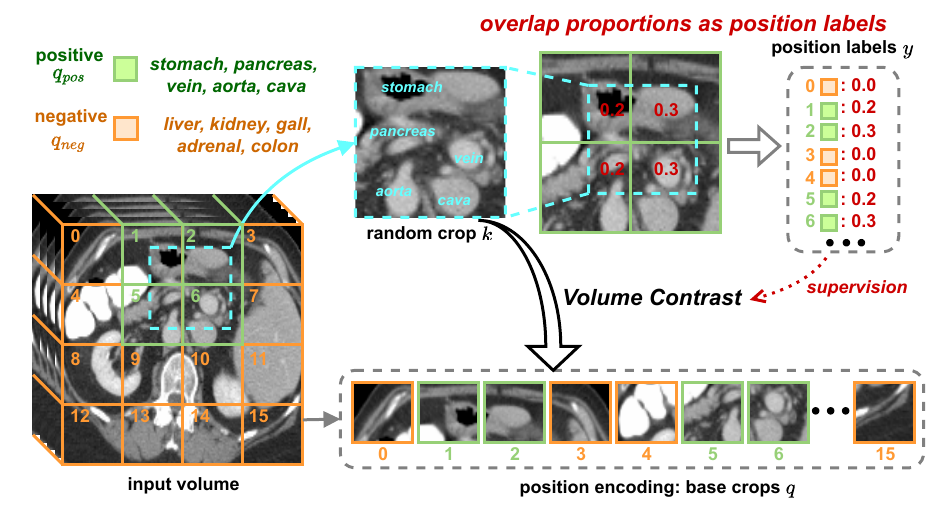}
	\caption{\textbf{Generate position labels for supervision}. A pair of random crop $k$ and base crop $q$ are assigned as \emph{positive} if they share overlap areas, otherwise as \emph{negative}. We calculate the overlap proportions as position labels $y$, \emph{e.g.}, ${y}_{1},{y}_{2},{y}_{5},{y}_{6}$ are assigned as $0.2,0.3,0.2,0.3$, respectively.}
	\label{fig_position}
\end{figure}

\begin{figure*}
	\centering
	\includegraphics[width=1\linewidth]{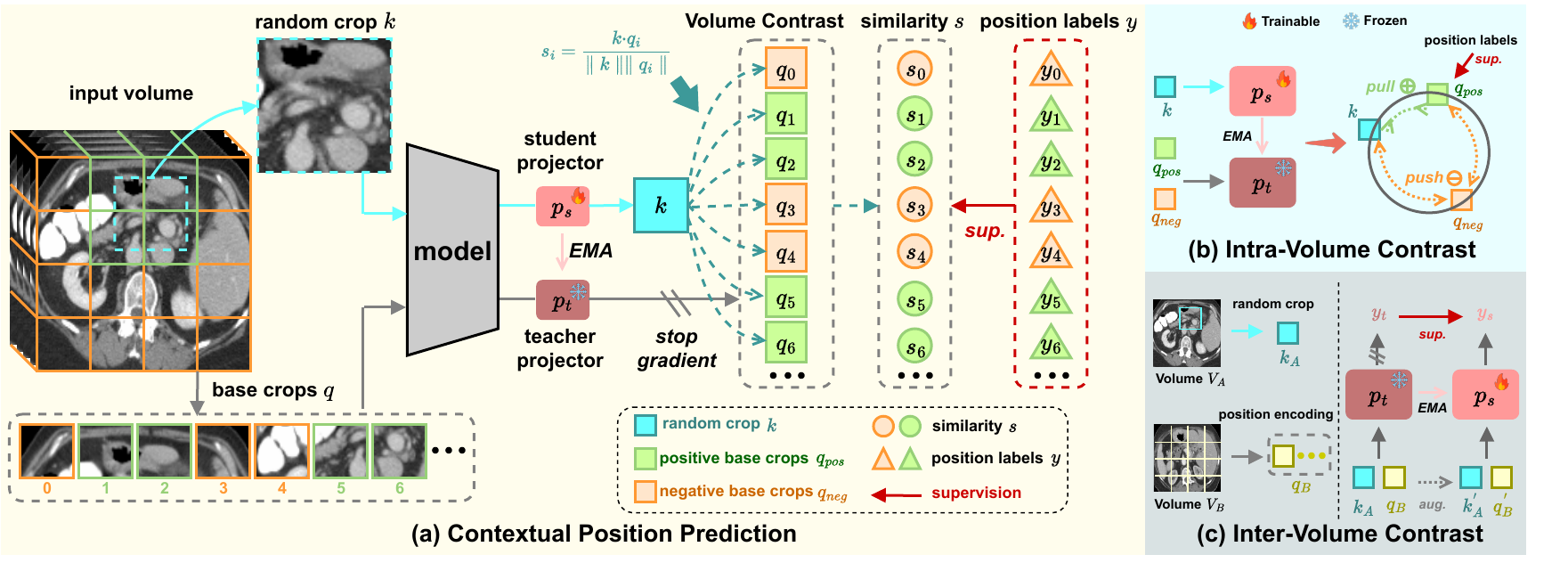}
	\caption{\textbf{Overall framework of VoCo}. \textbf{(a)} First, we generate base crops $q$ with corresponding position labels $y$ (Sec.~\ref{sec3_1} \& Fig.~\ref{fig_position}). Then we input the random crop $k$ and base crops $q$ for contextual position prediction. Specifically, we employ a student-teacher module to project $k$ and $q$ separately, where the teacher projector is frozen and updated from the student projector with Exponential Moving Average (EMA). Finally, we conduct volume contrast between $k$ and $q$ to predict similarity $s$ (Eq.~\ref{Eq_similarity}), where $s$ is supervised by position labels $y$ (Eq.~\ref{Eq_L_pred}). \textbf{(b)} We use the position labels to supervise the intra-volume contrast on $k$, $q_{pos}$, and $q_{neg}$, where $k$, $q_{pos}$, and $q_{neg}$ are from the same volume. \textbf{(c)} We extract random crop $k_{A}$ and base crops $q_{B}$ from different volumes $V_{A}$ and $V_{B}$ for inter-volume contrast.}
	\label{fig_framework}
\end{figure*}

\subsection{Generate Position Labels for Supervision}\label{sec3_1}

The pivotal procedure is to generate position labels for self-supervision. We propose to leverage the inherent geometric context priors in 3D medical images. As illustrated in Fig.~\ref{fig_position}, given an input volume $V$, we first randomly crop a sub-volume $k$, with the objective of constructing positive and negative pairs with $k$ for contrastive learning. Specifically, we propose to employ position encoding to generate $n$ non-overlap base crops ${q}_{i}, {i}{\in}{n}$. For example, $n=4{\times}4$ base crops are generated in Fig.~\ref{fig_position}, where each base crop ${q}_{i}$ represents a distinct region of the input volume.

Within human body anatomy, various organs are situated in distinct regions, leading to a potential way for us to form positive and negative pairs. As shown in Fig.~\ref{fig_position}, the random crop $k$ and the positive base crops ${q}_{pos}$ exhibit overlap areas, whereas the negative base crops ${q}_{neg}$, lacking such overlaps, more likely encompass different organs (not absolutely). For example, in Fig.~\ref{fig_position}, $k$ and ${q}_{pos}$ both contain \emph{stomach, pancreas, vein, aorta}, and \emph{cava}, while $k$ and ${q}_{neg}$ exhibit different organ information.
Thus, we can employ the position encoding to construct positive and negative pairs for contrastive learning. 

Previous contrastive learning methods~\cite{simclr,simsiam,moco,mocov3} mainly employ InfoNCE loss~\cite{infonce} to maximize the mutual information of positive pairs. In this paper, we propose to generate labels with specific values to supervise the correlation extent of positive pairs, \emph{i.e.}, with labels to reflect \textbf{how similar between \bm{$k$} and \bm{${q}_{pos}$}}. 
It can be observed that the correlation between $k$ and ${q}_{pos}$ is associated with their overlap proportions. Intuitively, if a positive base crop ${q}_{pos}$ shares more overlap areas with $k$, this ${q}_{pos}$ would be more similar with $k$. Thus, as shown in Fig.~\ref{fig_position}, we propose to assign the overlap proportions as the values of position labels $y$, enabling us to measure the similarity between $k$ and ${q}_{pos}$. In contrast, the position labels ${y}$ of ${q}_{neg}$ are assigned to $0$. In this way, we leverage the overlap proportions between $k$ and $q$ to supervise the contextual position prediction results. 

\subsection{Volume Contrast for Contextual Position Prediction}

The overall framework of VoCo is present in Fig.~\ref{fig_framework}. Specifically, we propose a novel pretext task, \emph{i.e.}, contextual position prediction, which employs volume contrast to predict the contextual positions of a random crop $k$. This pretext task includes: \textbf{(1)} intra-volume contrast among $k$, $q_{pos}$, and $q_{neg}$, where $k$, $q_{pos}$, and $q_{neg}$ are from the same volume; \textbf{(2)} inter-volume contrast between different volumes $V_{A}$ and $V_{B}$, which is established by consistency regularization with a typical student-teacher module~\cite{simclr,simsiam,moco,mocov3}.

\textbf{Contextual position prediction}. As shown in Fig.~\ref{fig_framework}(a), given an input volume, we first extract a random crop $k$ and a group of base crops $q$, where the corresponding position labels $y_{i}$ for $q_{i}$ are generated as Sec.~\ref{sec3_1} and Fig.~\ref{fig_position}. Then we feed $k$ and $q$ into the model to extract high-dimension features. After extracting the features, we employ a typical momentum-based student-teacher module~\cite{moco,mocov3} to project $k$ and $q$ separately. Specifically, the teacher projector $p_{t}$ is frozen during training, where its parameters ${\theta}_{t}$ are updated from the parameters ${\theta}_{s}$ of the student projector $p_{s}$ by Exponential Moving Average (EMA):
\begin{equation}\label{Eq_ema}
	{\theta}_{t} = {\rho}{\theta}_{t} + (1-{\rho}){\theta}_{s},
\end{equation}
where ${\rho}$ is the momentum factor and is empirically set to $0.9$. The momentum-based student-teacher module is effective in contrastive learning~\cite{moco,mocov3}, which enables stable training and avoids feature collapse~\cite{simclr,simsiam,swav}.

With features extracted from the projectors, we conduct 3D adaptive average pooling to resize $k$ and $q$ as one dimension features, \emph{i.e.}, $k{\in}{\mathbb{R}}^{1{\times}{C}}$ and $q{\in}{\mathbb{R}}^{1{\times}{C}}$, where $C$ is the number of feature dimensions. Then, we calculate the similarity $s_{i}$ between random crop $k$ and each base crop ${q}_{i}$. Specifically, we use cosine similarity to compute $s_{i}$ as:
\begin{equation}\label{Eq_similarity}
	s_{i}=CosSim(k, {q}_{i})=\frac{k{\cdot}{{q}_{i}}}{\parallel k\parallel \parallel {q}_{i} \parallel}, i{\in}n,
\end{equation}
where $s_{i}$ denotes the similarity between $k$ and ${q}_{i}$, which ranges from $0$ to $1$.

Intuitively, higher $s_{i}$ represents that $k$ has higher probabilities to share overlap proportions with ${q}_{i}$. In this way, we can predict the contextual position by calculating the similarity $s$. Then, we leverage the generated position labels $y$ to supervise the predicted similarity $s$. The formulation of prediction loss function $L_{pred}$ is based on entropy. Specifically, we first calculate the distance $d$ between similarity $s$ and position labels $y$:
\begin{equation}\label{Eq_dist}
	d_{i}=|{y}_{i}-{s}_{i}|, i{\in}n,
\end{equation}
where $|.|$ denotes the absolute value. Note that both $s$ and $y$ are ranging from $0$ to $1$. Then, $L_{pred}$ is formulated as:
\begin{equation}\label{Eq_L_pred}
	L_{pred}=-\frac{1}{n}\sum_{i{\in}n}^{n}log(1-d_{i}).
\end{equation}
\textbf{Remark}. Although we assign the position labels $y$ as $0$ for all negative base crops $q_{neg}$, there are instances where the random crop $k$ may resemble some $q_{neg}$. Without labels during pre-training, constructing absolutely ideal negative pairs in contrastive learning remains challenging~\cite{supervised_CL,moco,mocov3}. Nevertheless, the overall distances among negative pairs remain substantial. Thus, following previous methods~\cite{moco,mocov3,infonce}, we adopt the average entropy of distances in Eq.~\ref{Eq_L_pred}.

\textbf{Intra-volume contrast}. As shown in Fig.~\ref{fig_framework}(b), we conduct intra-volume contrast on a triplet: random crop $k$, positive base crop $q_{pos}$, and negative base crop $q_{neg}$. Specifically, we pull $k$ and $q_{pos}$ closer, push $k$ and $q_{pos}$, $q_{pos}$ and $q_{neg}$ apart from each other. For random crop $k$, we use position labels $y$ to supervise the process of contrastive learning. For $q$, we design a regularization loss $L_{reg}$ to enforce the feature discrepancy between each pair of $q_{i}$ and $q_{j}$:
\begin{equation}\label{Eq_L_reg}
	L_{reg}=\frac{2}{n(n-1)}\sum_{i,j{\in}n,i{\neq}j}^{n}|s_{ij}|, i,j{\in}n, i{\neq}j,
\end{equation}
where $s_{ij}$ is the cosine similarity between different $q_{i}$ and $q_{j}$ as follows:
\begin{equation}\label{Eq_cos}
	s_{ij}=CosSim({q}_{i}, {q}_{j})=\frac{{q}_{i}{\cdot}{{q}_{j}}}{\parallel {q}_{i}\parallel \parallel {q}_{j} \parallel}, i,j{\in}n, i{\neq}j.
\end{equation}

\textbf{Inter-volume contrast}. As shown in Fig.~\ref{fig_framework}(c), we extract random crop $k_{A}$ from volume $V_{A}$ and base crops $q_{B}$ from volume $V_{B}$ to establish inter-volume contrast, where volumes $V_{A}$ and $V_{B}$ are sampled from the same batch during training. It is worth noting that $V_{A}$ and $V_{B}$ are sampled from the same anatomical region, \emph{e.g.}, both from the abdomen region or both from the chest region.

Specifically, we adopt a simple-yet-effective consistency regularization method~\cite{yang2023revisiting,moco,mocov3} for inter-volume contrast. We first employ feature augmentation (aug. in Fig.~\ref{fig_framework}) to $k_{A},q_{B}$ and get $k^{'}_{A},q^{'}_{B}$,  The augmentation here is a simple Dropout~\cite{dropout} as~\cite{yang2023revisiting}. 
Then we fed the features before and after augmentation into $p_{s}$ and $p_{t}$, respectively. After the projectors, we also compute the cosine similarity for contrastive learning: 
\begin{equation}\label{Eq_inter}
	y_{s}=CosSim({k}_{A},{q}_{B}), y_{t}=CosSim({k}^{'}_{A},{q}^{'}_{B}).
\end{equation}
Then we formulate the loss function $L_{inter}$ as:
\begin{equation}\label{Eq_L_inter}
	L_{inter}=-\frac{1}{n}\sum_{i{\in}n}^{n}log(1-|y_{s}-y_{t}|).
\end{equation}

\textbf{Overall loss function for SSL}. Overall, the loss function $L_{SSL}$ for SSL is the combination:
\begin{equation}\label{Eq_L_ssl}
	L_{SSL} = L_{pred} + L_{reg} + L_{inter},
\end{equation}
where we empirically set the same weights for three loss functions~\cite{VoCo}, since we consider their importance equally. 

\begin{figure}
	\centering
	\includegraphics[width=1\linewidth]{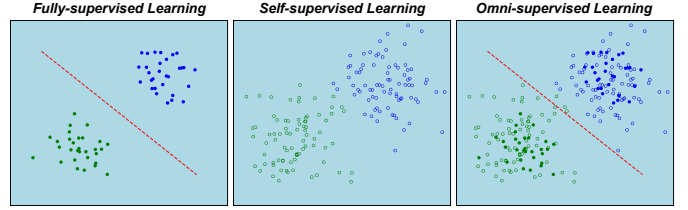}
	\caption{Differences among fully-, self-, and omni-supervised learning. Solid and hollow markers denote labeled and unlabeled data, respectively. Dashed lines denote decision boundaries between different classes.}
	\label{fig_decision}
\end{figure}

\subsection{Towards Omni-supervised Pre-training}

\begin{algorithm}
\caption{Omni-supervised Pre-training}\label{omni_algorithm}
\KwData{Labeled segmentation data: $\bm{(X_{L},Y_{L})}$. Unlabeled data: $\bm{X_{U}}$}
\KwResult{Pre-trained model $\bm{M}$}

\textbf{First stage:}\\

\mbox{}\quad {Fully-supervised training $\bm{M}$ $\leftarrow$ $\bm{(X_{L},Y_{L})}$}\;

\mbox{}\quad {Self-supervised training $\bm{M}$ $\leftarrow$ $\bm{X_{U}}$} [$L_{SSL}$ Eq.~\ref{Eq_L_ssl}]\;

\textbf{Second stage:}\\

\mbox{}\quad {Generate pseudo labels $\bm{Y_{U}}$ $\leftarrow$ $\bm{(M, X_{U})}$}\;

\mbox{}\quad {Semi-supervised training $\bm{M}$ $\leftarrow$ $\bm{(X_{L}, Y_{L}, X_{U}, Y_{U})}$}\;

\mbox{}\quad {Self-supervised training $\bm{M}$ $\leftarrow$ $\bm{X_{U}}$} [$L_{SSL}$ Eq.~\ref{Eq_L_ssl}]\;

\end{algorithm}

\begin{table*}
	\setlength{\abovecaptionskip}{0.pt}
	\setlength{\belowcaptionskip}{-0.em}
	\centering
	\footnotesize
 \caption{\textbf{PreCT-160K} contains \textbf{160K} CT from 30 public datasets, with more than \textbf{42M} slices covering the anatomical structures. 10K is used in our preliminary study~\cite{VoCo}.}
\begin{threeparttable}
	\begin{tabular}{ccccc}
		\toprule[1.2pt]
		\multirow{2}{*}{\textbf{Dataset}} &\multirow{2}{*}{\textbf{Anatomical Region}} &\multicolumn{2}{c}{\textbf{Pre-training Scale}} &\multirow{2}{*}{\textbf{Number of Volumes}}\\
        \cline{3-4}
        & &\textbf{10K} &\textbf{160K} &\\
		\hline
        BTCV~\cite{btcv} &Abdomen &\CheckmarkBold &\CheckmarkBold &24\\
        TCIA-Covid19~\cite{tcia} &Chest &\CheckmarkBold &\CheckmarkBold &722 \\
        LUNA16~\cite{luna} &Chest &\CheckmarkBold &\CheckmarkBold &843 \\
        FLARE23~\cite{FLARE22} &Abdomen &\CheckmarkBold &\CheckmarkBold &4000 \\
        HNSCC~\cite{HNSCC} &Head/Neck  &\CheckmarkBold &\CheckmarkBold  &1071 \\
        STOIC 2021~\cite{stoic} &Chest &\CheckmarkBold &\CheckmarkBold &2000 \\
        LIDC~\cite{LIDC} &Chest &\CheckmarkBold &\CheckmarkBold &1018 \\
        TotalSegmentator~\cite{Total} &104 
        Anatomic Structures &\CheckmarkBold &\CheckmarkBold &1203 \\
        Tumor datasets~\cite{MSD,kits,chaos,panc_ct,abdomenct1k,Difftumor} &Abdomen & &\CheckmarkBold &1334 \\
        WORD~\cite{word} &Abdomen & &\CheckmarkBold &120\\
        AMOS22~\cite{amos} &Abdomen & &\CheckmarkBold &300\\
        DeepLesion~\cite{DeepLesion} &Abdomen & &\CheckmarkBold &1618\\
        PANORAMA~\cite{PANORAMA} &Abdomen & &\CheckmarkBold &2238\\
        AbdomenAtlas1.0~\cite{atlas} &Abdomen & &\CheckmarkBold &5195 \\
        OPC-Radiomics~\cite{OPC} &Oropharyngeal & &\CheckmarkBold &606\\
        HeadNeckCT~\cite{HeadNeckPET} &Head/Neck & &\CheckmarkBold &504 \\
        Qin-Headneck~\cite{QIN-HEADNECK} &Head/Neck & &\CheckmarkBold &892 \\
        TCGA-HNSC~\cite{TCGA-HNSC} &Head/Neck & &\CheckmarkBold &1274 \\
        CT COLONOGRAPHY~\cite{CT-COLONOGRAPHY} &Chest, Abdomen, Colon cancer & &\CheckmarkBold  &1730 \\
        MELA~\cite{MELA} &Chest & &\CheckmarkBold &770\\
        StonyBrookChestCT~\cite{StonyBrookChestCT} &Chest & &\CheckmarkBold &2316\\
        CT-RATE~\cite{CT-CLIP} &Chest & &\CheckmarkBold &47149\\
        NLST~\cite{NLST} &Chest & &\CheckmarkBold &84830 \\
        \rowcolor{pink}
        \textbf{Total} & & & &\textbf{160167}\\
        \toprule[1.2pt]
	\end{tabular}
    \end{threeparttable}        
\label{table_dataset}
\end{table*}

As shown in Fig.~\ref{fig_decision}, both fully- and self-supervised learning have specific merits and drawbacks. \textbf{(a)} Fully-supervised learning can learn discriminative decision boundaries with the guidance of labels yet it is constrained by the lack of labeled data. \textbf{(b)} SSL can leverage large-scale unlabeled data. However, lacking annotations for supervision, it struggles with learning clear decision boundaries between distinct classes.
To this end, we propose omni-supervised pre-training to effectively leverage both labeled and unlabeled data, as described in Algorithm~\ref{omni_algorithm}. Our omni-supervised learning amalgamates the strengths of both fully- and self-supervised learning and effectively unleashes the potential of labeled and unlabeled data. 

\textbf{Curating labeled segmentation dataset $\bm{X_{L}},\bm{Y_{L}}$}. The PreCT-160K dataset includes extensive labeled segmentation datasets. However, many of these datasets have incomplete labels~\cite{xie2023learning,abdomenct1k,atlas}, such as one dataset containing solely liver labels and another with only pancreas labels. Thus, we first ensemble various models to generate complete labels $Y_{L}$ for $X_{L}$ and curate a small subset of labeled data from PreCT-160K. This subset, which we named VoComni, contains 20K volumes spanning 20 different organ and tumor classes, which will be released with PreCT-160K for fostering future research. All the val and test sets are unseen in PreCT-160K and VoComni.

\textbf{Semi-supervised learning is a scalable learner}. To effectively leverage labeled and unlabeled data, we propose to conduct semi-supervised learning~\cite{CISC_R,yang2022st++,yang2023revisiting} to borrow the knowledge from labeled data to large-scale unlabeled data. Notably, segmentation emerges as a pivotal technique in supervised training~\cite{SuPreM,clipdriven,dodnet}, given that many medical tasks demand a granular understanding at the pixel level for accurate diagnosis. Previous works~\cite{wang2024towards,dodnet,wu2022minimizing} only leveraged a few hundred cases for semi-supervised segmentation. However, complex designs of semi-supervised segmentation will significantly increase the burden of training, which is not feasible to our large-scale data. In this paper, we adopt a simple semi-supervised learning baseline and scale up the data to 160K volumes.
We find that incorporated with VoCo, the simplest semi-supervised baseline can already achieve competitive results. 
As shown in Algorithm~\ref{omni_algorithm}, we first curate a labeled segmentation dataset $(X_{L},Y_{L})$ from PreCT-160K and perform supervised segmentation training in the first stage. Then in the second stage, we generate pseudo labels $Y_{U}$ for unlabeled data $X_{U}$, aiming to perform semi-supervised learning on both $X_{L}$ and $X_{U}$. \textbf{Note that} SSL is collaboratively integrated with semi-supervised training in both two stages. In this way, we amalgamate the strengths of self- and semi-supervised learning, advancing towards omni-supervised pre-training.

\begin{table}
	\setlength{\abovecaptionskip}{0.pt}
	\setlength{\belowcaptionskip}{-0.em}
	\centering
	\footnotesize
\caption{\textbf{48 downstream datasets in our benchmark}. \textbf{28} of them are unseen in pre-training (denoted with $\dagger$). \textbf{18} of them are with less than 50 cases for finetuning (denoted with $^*$). The labels of val sets are unseen in pre-training. Test sets are evaluated on public leaderboards (if available).}
\begin{threeparttable}
	\begin{tabular}{ccc}
		\toprule[1.2pt]
		\textbf{Dataset} &\textbf{Modality} &\textbf{Task}\\
		\hline
        BTCV$^*$~\cite{btcv} &CT &Abdomen Seg.\\
        AMOS22~\cite{amos} &CT &Abdomen Seg.\\
        WORD~\cite{word} &CT &Abdomen Seg.\\
        FLARE22$^*$~\cite{FLARE22} &CT &Abdomen Seg. \\
        FLARE23~\cite{FLARE22} &CT &Abdomen Seg. \\
        Abdomenct1k~\cite{abdomenct1k} &CT &Abdomen Seg.\\
        AbdomenAtlas~\cite{atlas} &CT &Abdomen Seg.\\
        TotalSegmentator~\cite{Total} &CT &104 Structures Seg.\\
        MM-WHS$\dagger$$^*$~\cite{mmwhs} &CT &Heart Seg.\\
        % ASOCA$\dagger$$^*$~\cite{ASOCA} &CT &Coronary Seg.\\
        AVT$\dagger$$^*$~\cite{avt} &CT &Aorta Seg.\\
        CHAOS$^*$~\cite{chaos} &CT &Liver Seg.\\
        Sliver07$\dagger$$^*$~\cite{SLIVER07} &CT &Liver Seg.\\
        IRCADb$\dagger$$^*$~\cite{3D-IRCADb} &CT &Liver Tumor Seg.\\
        KiTS~\cite{kits} &CT &Kidney Tumor Seg.\\
        KiPA22$\dagger$~\cite{Kipa} &CT &Kidney Tumor Seg.\\
        TCIA-Panc.$^*$~\cite{panc_ct} &CT &Panc. Seg.\\
        PANORAMA~\cite{PANORAMA} &CT &Panc. Tumor Seg.\\
        SegThor$\dagger$$^*$~\cite{segthor} &CT &Thoracic Risk Seg.\\
        BHSD$\dagger$~\cite{BHSD} &CT &Brain Bleed Seg.\\
        StructSeg19$\dagger$$^*$~\cite{StructSeg} &CT &Nasopharynx Cancer Seg.\\
        Verse20$\dagger$~\cite{verse} &CT &Vertebrae Seg.\\
        Covid-19-20$\dagger$~\cite{covid} &CT &Covid Seg.\\
        FUMPE$\dagger$$^*$~\cite{FUMPE} &CT &Pulmonary Embolism Seg.\\
        Parse22$\dagger$~\cite{Parse22} &CT &Pulmonary Artery Seg.\\
        AIIB23$\dagger$~\cite{AIIB23} &CT &Fibrotic Lung Seg.\\
        CC-CCII$\dagger$~\cite{CC-CCII} &CT &Covid Classi.\\
        LUNA16~\cite{luna} &CT &Lung Nodule Classi.\\
        AutoPET-II23$\dagger$~\cite{AutoPET} &PET-CT &HeadNeck Lesion Seg.\\
        % PICAI~\cite{picai} &MRI &Prostate Cancer Seg.\\
        AMOS-MRI$\dagger$$^*$~\cite{amos} &MRI &Abdomen Seg.\\
        MM-WHS-MRI$\dagger$$^*$~\cite{mmwhs} &MRI &Heart Seg.\\
        ACDC$\dagger$~\cite{ACDC} &MRI &Heart Seg.\\
        ATLAS-MRI$\dagger$$^*$~\cite{ATLAS-MRI} &MRI &Liver Tumor Seg.\\
        BraTs21$\dagger$~\cite{brats} &MRI &Brain Tumor Seg.\\
        IXI$\dagger$~\cite{cyclemorph} &MRI &Brain MRI Registration\\
        OASIS$\dagger$~\cite{marcus2007open} &MRI &Brain MRI Registration\\
        CTRG-Chest$\dagger$~\cite{CTRG} &VLP &Report Generation\\
        CT-RATE~\cite{CT-CLIP} &VLP &Vocabulary Classi.\\
        CT-RATE~\cite{CT-CLIP} &VLP &Report-Volume Retrieval\\
        \hline
        \emph{MSD Challenge}~\cite{MSD} & &\\
        Task01 Brain$\dagger$ &MRI &Brain Tumor Seg.\\
        Task02 Heart$\dagger$$^*$ &MRI &Heart Seg.\\
        Task03 Liver &CT &Liver Tumor Seg.\\
        Task04 Hip.$\dagger$ &MRI &Hip. Seg.\\
        Task05 Pros.$\dagger$$^*$ &MRI &Prostate Seg.\\
        Task06 Lung$^*$ &CT &Lung Cancer Seg.\\
        Task07 Panc. &CT &Pancreas Tumor Seg.\\
        Task08 Vessel$\dagger$ &CT &Vessel Tumor Seg.\\
        Task09 Spleen$^*$ &CT &Spleen Seg. \\
        Task10 Colon &CT &Colon Cancer Seg.\\
        \toprule[1.2pt]
	\end{tabular}
    \end{threeparttable}        
\label{table_downstream_dataset}
\end{table}%

\begin{table}
	\setlength{\abovecaptionskip}{0.pt}
	\setlength{\belowcaptionskip}{-0.em}
	\centering
	\footnotesize
\caption{\textbf{Our benchmark is more comprehensive}, with more tasks and data for evaluation. \textbf{Eval Sets} denote the number of downstream datasets.}
\begin{threeparttable}
	\begin{tabular}{lccccc}
		\toprule[1.2pt]
		\multirow{2}{*}{\textbf{Method~(Publication)}} &\multicolumn{4}{c}{\textbf{Downstream Tasks}} &\multirow{2}{*}{\textbf{Eval Sets}}\\
        \cline{2-5}
        &Seg. &Cls. &Reg. &VL &\\
		\hline
        PCRL~\cite{PCRLv2}~(TPAMI23) &\CheckmarkBold &\CheckmarkBold & & &4\\
        GVSL~\cite{geo}(CVPR23) &\CheckmarkBold &\CheckmarkBold & & &4\\
        Swin.~\cite{swin}~(CVPR23) &\CheckmarkBold & & & &11\\
        Alice~\cite{Alice}~(ICCV23) &\CheckmarkBold & & & &3\\
        Univer.~\cite{clipdriven}~(ICCV23) &\CheckmarkBold & & & &14\\
        Unim.~\cite{UniMiSS+}~(TPAMI24) &\CheckmarkBold &\CheckmarkBold & & &10\\ 
        MedC.~\cite{continual}~(CVPR24) &\CheckmarkBold & & & &9\\
        VoCo-v1~\cite{VoCo}~(CVPR24) &\CheckmarkBold &\CheckmarkBold & & &6\\
        SuPrem~\cite{SuPreM}~(ICLR24) &\CheckmarkBold & & & &3\\
        \rowcolor{mygray}
        \textbf{VoCo} &\CheckmarkBold &\CheckmarkBold &\CheckmarkBold &\CheckmarkBold &\textbf{48}\\
        \toprule[1.2pt]
	\end{tabular}
    \end{threeparttable}        
\label{table_downstream_compare}
\end{table}%

\section{Experiments}

\subsection{Dataset and Implementation Details}

\indent \textbf{Pre-training dataset}\footnote{\href{https://huggingface.co/datasets/Luffy503/PreCT-160K}{https://huggingface.co/datasets/Luffy503/PreCT-160K}}. In this paper, we curate the existing largest dataset medical image pre-training PreCT-160K, as shown in Table~\ref{table_dataset}. PreCT-160K is collected from diverse sources and underwent thorough pre-processing to ensure a consistent data format for training. Specifically, to address variations in sizes, spacing, and intensity across volumes from different anatomical regions, we have devised tailored pre-processing protocols. Since in PreCT-160K, data from chest regions cover larger proportions, we simply balance the sampling during pre-training.
For VoComni dataset\footnote{\href{https://huggingface.co/datasets/Luffy503/VoComni}{https://huggingface.co/datasets/Luffy503/VoComni}}, we use model ensembling to generate pseudo labels, where we discard the volumes with low prediction confidence. Consequently, we have created a segmentation dataset comprising 20K pseudo-labeled volumes (encompassing 20 organ and tumor classes) for our omni-supervised pre-training.

% The description and the pre-processing details are introduced in the supplementary materials.

\textbf{Evaluation benchmark}. We build a large-scale evaluation benchmark as shown in Table~\ref{table_downstream_dataset}, which includes 48 downstream datasets for various tasks. It can seen in Table~\ref{table_downstream_compare} that our evaluation benchmark is more comprehensive than that of previous works~\cite{PCRLv2,geo,swin,Alice,UniMiSS+,continual,SuPreM}. A number of datasets~\cite{amos,FLARE22,kits,abdomenct1k} are evaluated on the public leaderboards. If the test sets and public leaderboards are not available, we report the offline val sets results with the same data splits for fair comparisons.

\textbf{Experiment settings}. We first conduct pre-training on PreCT-160K then finetune the pre-trained models on 48 downstream datasets (Table~\ref{table_downstream_dataset}) for evaluation. We adopt both SwinUNETR~\cite{swinunetr} and nnUNet~\cite{nnunet} as the backbones for pre-training. Specifically, Swin-Base (B), Swin-Large (L), and Swin-Huge (H) are all adopted for training, with feature sizes of $48$, $96$, and $192$ in SwinUNETR~\cite{swinunetr}, respectively. This project is supported by NVIDIA SuperPOD hardware. 8 $\times$ NVIDIA H800 GPUs are used for pre-training and all the downstream tasks can be done with one H800 or 3090 GPU. It spent over 10,000 GPU hours in downstream evaluation. Our implementation codes are all open-source and support both MONAI~\cite{Monai} and nnUNet~\cite{nnunet} frameworks.

\begin{table*}
	\setlength{\abovecaptionskip}{0.pt}
	\setlength{\belowcaptionskip}{-0.em}
	\centering
	\footnotesize
 \caption{The DSC (\%) of seven widely-used segmentation datasets, \emph{i.e.}, BTCV\cite{btcv}, AMOS22\cite{amos}, WORD\cite{word}, FLARE22\cite{FLARE22}, FLARE23\cite{FLARE22}, TotalSegmentator\cite{Total}, and AbdomenAtlas\cite{atlas}. The state-of-the-art results among previous methods are \underline{underlined} while the best results are \textbf{bolded}. Note that \cite{stunet,clipdriven,SuPreM} are fully-supervised pre-training methods. Since \cite{FLARE22,Total,atlas} require huge computation costs, we only report the results of advanced methods for comparisons. 
 Compared with models without pre-training, VoCo pre-training brings average \textcolor{red}{\bm{$+3.12\%$}} DSC improvements. }

\begin{threeparttable}
	\begin{tabular}{ccc|ccccccc|c}
		\toprule[1.2pt]
		\textbf{Method} &\textbf{Model~(Params)} &\textbf{Data Scale} &\textbf{BTCV} &\textbf{AMOS} &\textbf{WORD} &\textbf{FLA22} &\textbf{FLA23} &\textbf{Total}  &\textbf{Atlas} &\textbf{\textcolor{cyan}{\bm{$\triangle$}$\%$(AVG)}}\\
        \hline
        \multirow{5}{*}{\textbf{From Scratch}} &3D-UNet~(19M) & &80.98 &84.02 &83.21 &87.58 &- &- &-\\
        &UNETR~\cite{unetr}~(115M) & &79.82 &82.52 &79.77 &89.02 &- &- &-\\
        &nnUNet~\cite{nnunet}~(31M) & &79.29 &88.79 &86.04 &91.38 &\underline{91.54} &82.26 &88.77\\
        &Swin-B~\cite{swinunetr}~(72M) & &82.79 &87.19 &84.56 &89.18 &89.72 &80.97 &88.12\\
        &Swin-L~\cite{swinunetr}~(290M) & &83.52 &87.53 &83.17 &89.49 &89.33 &82.04 &88.56\\
        &Swin-H~\cite{swinunetr}~(1.2B) & &79.36 &86.14 &82.46 &87.96 &89.71 &82.78 &88.97\\
        \hline
        \textcolor{cyan}{\textbf{\emph{General Pre-training}}} & & & & & & & & &\\
        MAE3D~\cite{MAE,MAE3D,mim} &UNETR &10k &82.48 &82.71 &74.27 &89.31 &- &- &-\\
        MoCo v3~\cite{moco,mocov3} &Swin-B &1.6k &79.54 &80.95 &71.16 &83.22 &- &- &-\\
        \hline
        \textcolor{cyan}{\textbf{\emph{Medical Pre-training}}} & & & & & & & & &\\
        MG~\cite{Modelgen} &3D-UNet &0.6k &81.45 &81.27 &85.50 &85.02 &- &- &-\\
        DoDNet~\cite{dodnet} &3D-UNet &1k &81.10 &79.63 &85.90 &86.19 &- &- &-\\
        Unimiss~\cite{unimiss,UniMiSS+} &MiT(48M) &5k &82.05 &86.26 &83.37 &89.17 &- &- &-\\
        % MiM~\cite{mim} &Swin-B &10k &\underline{85.61} &- &- &89.67 &- &- &-\\
        SwinUNETR~\cite{swin} &Swin-B &5k &82.58 &85.68 &84.88 &89.31 &- &- &-\\
        % STUNet~\cite{stunet} &STUNet-H~(1.4B) &1.2k &83.83 &83.11 &77.42 &89.87 &- &- &-\\
        Universal Model~\cite{clipdriven} &Swin-B &2.1k &83.74 &88.01 &85.19 &91.11 &- &- &-\\
        SuPreM~\cite{SuPreM} &Swin-B &2.1k &\underline{85.32} &\underline{88.14} &85.97 &\underline{91.37} &89.98 &\underline{82.96} &\underline{89.16}\\
        VoCo-v1~\cite{VoCo} &Swin-B &10k &84.51 &88.06 &\underline{86.11} &91.29 &90.07 &80.46 &89.13\\
        % \rowcolor{mygray}
        % VoCo &nnUNet &160k &82.37 &\textbf{90.89} &86.52 &91.51 &91.01 &84.90 &88.89 &\textcolor{cyan}{$\uparrow$$0.86$}\\
        \rowcolor{mygray}
        VoCo &Swin-B &160k &\textbf{86.64} &\textbf{90.86} &\textbf{86.88} &92.17 &90.34 &84.84 &90.38 &\textcolor{cyan}{$\uparrow$$2.79$}\\
        \rowcolor{mygray}
        VoCo &Swin-L &160k &86.05 &89.43 &86.77 &92.37 &91.56 &85.27 &90.90 &\textcolor{cyan}{$\uparrow$$2.67$}\\
        \rowcolor{mygray}
        VoCo &Swin-H &160k &86.21 &88.79 &86.12 &\textbf{92.65} &\textbf{92.30} &\textbf{86.18} &\textbf{91.41} &\textcolor{cyan}{$\uparrow$$3.75$}\\
        \toprule[1.2pt]
	\end{tabular}
    \end{threeparttable}        
\label{table_abdomen_performance}
\end{table*}

\begin{table*}
	\setlength{\abovecaptionskip}{0.pt}
	\setlength{\belowcaptionskip}{-0.em}
	\centering
	\footnotesize
 \caption{The DSC (\%) of \textbf{24} downstream segmentation tasks. nnUNet~\cite{nnunet} and Swin-B~\cite{swinunetr} are from scratch, others are with pre-training. 
 \textbf{Note that} the improvements are more significant for the datasets with fewer cases for finetuning (refer to Table~\ref{table_downstream_dataset}).
 Compared with models without pre-training, VoCo pre-training brings average \textcolor{red}{\bm{$+4.42\%$}} DSC improvements.}
\begin{threeparttable}
	\begin{tabular}{c|cccccccc}
		\toprule[1.2pt]
		\textbf{Method} &\textbf{Ab1k}~\cite{abdomenct1k} &\textbf{WHS}~\cite{mmwhs} &\textbf{AVT}~\cite{avt} &\textbf{CHAOS}~\cite{chaos} &\textbf{Sliver.}~\cite{SLIVER07}
        &\textbf{IR.}~\cite{3D-IRCADb} &\textbf{KiTs}~\cite{kits}
        &\textbf{Kipa.}~\cite{Kipa}
        \\
        \hline
        nnUNet~\cite{nnunet} &85.74 &88.72 &50.19 &94.53 &94.87 &51.26 &78.92 &88.99\\
        Swin-B~\cite{swinunetr} &85.76 &89.11 &46.76 &94.10 &94.96 &57.19 &78.61 &85.18\\
        \hline
        SwinUNETR~\cite{swin} &86.32 &89.06 &46.18 &94.98 &94.67 &55.69 &76.82 &85.14\\
        SuPrem~\cite{SuPreM} &86.40 &90.88 &58.85 &96.42 &96.72 &68.48 &78.38 &85.76\\
        \rowcolor{mygray}
        VoCo (nnUNet) &86.75 &89.53 &58.23 &96.01 &95.98 &60.84 &80.80 &\textbf{90.31}\\
        \rowcolor{mygray}
        VoCo (Swin-B) &\textbf{87.77} &\textbf{91.22} &\textbf{69.64} &\textbf{96.68} &\textbf{97.75} &\textbf{74.27} &\textbf{80.81} &87.54\\
        \hline
        \textcolor{cyan}{\bm{$\triangle$}(nnUNet)} &\textcolor{cyan}{$\uparrow$$1.01$}
        &\textcolor{cyan}{$\uparrow$$0.81$}
        &\textcolor{cyan}{$\uparrow$$8.04$}
        &\textcolor{cyan}{$\uparrow$$1.48$}
        &\textcolor{cyan}{$\uparrow$$1.11$}
        &\textcolor{cyan}{$\uparrow$$9.58$}
        &\textcolor{cyan}{$\uparrow$$1.88$}
        &\textcolor{cyan}{$\uparrow$$1.32$}\\
        \textcolor{cyan}{\bm{$\triangle$}(Swin-B)} &\textcolor{cyan}{$\uparrow$$2.01$}
        &\textcolor{cyan}{$\uparrow$$2.11$}
        &\textcolor{cyan}{$\uparrow$$22.88$}
        &\textcolor{cyan}{$\uparrow$$2.58$}
        &\textcolor{cyan}{$\uparrow$$1.79$}
        &\textcolor{cyan}{$\uparrow$$17.08$}
        &\textcolor{cyan}{$\uparrow$$2.20$}
        &\textcolor{cyan}{$\uparrow$$2.36$}\\
        
        \toprule[1.2pt]
		\textbf{Method}  &\textbf{Panc.}~\cite{panc_ct} &\textbf{PANO.}~\cite{chaos} &\textbf{Segthor}~\cite{segthor} &\textbf{BHSD}~\cite{BHSD} &\textbf{Struct.}~\cite{StructSeg} &\textbf{Verse.}~\cite{verse}
        &\textbf{Covid.}~\cite{covid}
        &\textbf{FUMPE}~\cite{FUMPE}
        \\
        \hline
        nnUNet~\cite{nnunet} &84.68 &78.06 &88.15 &35.02 &70.60 &65.13 &62.42 &48.62\\
        Swin-B~\cite{swinunetr} &84.38 &78.40 &87.90 &36.40 &\textbf{76.42} &62.01 &63.91 &50.31\\
        \hline
        SwinUNETR~\cite{swin} &84.53 &78.34 &87.23 &35.97 &53.36 &87.33 &65.90 &51.72\\
        SuPrem~\cite{SuPreM} &85.19 &79.92 &89.70 &32.82 &59.85 &89.54 &63.29 &51.98\\
        \rowcolor{mygray}
        VoCo (nnUNet) &\textbf{87.59} &79.52 &88.82 &37.04 &72.74 &\textbf{67.82} &65.35 &49.50\\
        \rowcolor{mygray}
         VoCo (Swin-B) &86.57 &\textbf{80.13} &\textbf{90.17} &\textbf{38.38} &75.58 &63.72 &\textbf{68.72} &\textbf{55.32}\\
         \hline
         \textcolor{cyan}{\bm{$\triangle$}(nnUNet)} &\textcolor{cyan}{$\uparrow$$2.91$}
        &\textcolor{cyan}{$\uparrow$$1.48$}
        &\textcolor{cyan}{$\uparrow$$1.46$}
        &\textcolor{cyan}{$\uparrow$$2.02$}
        &\textcolor{cyan}{$\uparrow$$2.15$}
         &\textcolor{cyan}{$\uparrow$$2.69$}
        &\textcolor{cyan}{$\uparrow$$2.93$}
        &\textcolor{cyan}{$\uparrow$$0.89$}\\
        \textcolor{cyan}{\bm{$\triangle$}(Swin-B)} &\textcolor{cyan}{$\uparrow$$2.19$}
        &\textcolor{cyan}{$\uparrow$$1.73$}
        &\textcolor{cyan}{$\uparrow$$2.38$}
        &\textcolor{cyan}{$\uparrow$$1.98$}
        &\textcolor{dark-red}{$\downarrow$$1.14$}
        &\textcolor{cyan}{$\uparrow$$1.71$}
        &\textcolor{cyan}{$\uparrow$$4.82$}
        &\textcolor{cyan}{$\uparrow$$5.01$}\\
        
        \toprule[1.2pt]
		\textbf{Method}  &\textbf{Parse.}~\cite{Parse22} &\textbf{AIIB.}~\cite{AIIB23} &\textbf{Auto.}~\cite{AutoPET} &\textbf{AM-MR.}~\cite{amos} &\textbf{WHS-MR}~\cite{mmwhs}
        &\textbf{ACDC}~\cite{ACDC}
        &\textbf{At-MR}~\cite{ATLAS-MRI} &\textbf{BraTs.}~\cite{brats} 
        \\
        \hline
        nnUNet~\cite{nnunet} &80.55 &88.72 &35.84 &72.56 &85.36 &92.12 &63.23 &\textbf{91.02}\\
        Swin-B~\cite{swinunetr} &82.78 &89.09 &25.25 &72.46 &86.13 &87.22 &60.40 &89.05\\
        \hline
        SwinUNETR~\cite{swin} &81.66 &89.05 &22.09 &72.89 &86.29 &89.47 &60.51 &87.33\\
        SuPrem~\cite{SuPreM} &82.88 &89.96 &24.68 &75.69 &85.79 &89.10 &64.64 &89.54\\
        \rowcolor{mygray}
        VoCo (nnUNet) &81.60 &90.12 &33.02 &74.38 &86.26 &\textbf{92.41} &68.19 &90.51\\
        \rowcolor{mygray}
         VoCo (Swin-B) &\textbf{83.87} &\textbf{90.44} &32.61 &\textbf{79.24} &\textbf{87.71} &89.51 &\textbf{69.80} &90.23\\
         \hline
         \textcolor{cyan}{\bm{$\triangle$}(nnUNet)} &\textcolor{cyan}{$\uparrow$$1.11$}
        &\textcolor{cyan}{$\uparrow$$1.41$}
        &\textcolor{dark-red}{$\downarrow$$2.82$}
        &\textcolor{cyan}{$\uparrow$$1.82$}
        &\textcolor{cyan}{$\uparrow$$0.90$}
        &\textcolor{cyan}{$\uparrow$$0.29$}
        &\textcolor{cyan}{$\uparrow$$4.96$}
        &\textcolor{dark-red}{$\downarrow$$0.51$}\\
        
        \textcolor{cyan}{\bm{$\triangle$}(Swin-B)} &\textcolor{cyan}{$\uparrow$$1.10$}
        &\textcolor{cyan}{$\uparrow$$1.35$}
        &\textcolor{cyan}{$\uparrow$$7.36$}
        &\textcolor{cyan}{$\uparrow$$6.78$}
        &\textcolor{cyan}{$\uparrow$$1.58$}
        &\textcolor{cyan}{$\uparrow$$2.28$}
        &\textcolor{cyan}{$\uparrow$$9.40$}
        &\textcolor{cyan}{$\uparrow$$1.18$}\\
        \toprule[1.2pt]
	\end{tabular}
    \end{threeparttable}        
\label{table_other_downstream_performance}
\end{table*}

\begin{table*}
	\setlength{\abovecaptionskip}{0.pt}
	\setlength{\belowcaptionskip}{-0.em}
	\centering
	\footnotesize
 \caption{The DSC (\%) of MSD 10-Task Challenge~\cite{MSD}. We report the results of the same folds defined by nnUNet~\cite{nnunet} for fair comparison. We report the tumor DSC for Task03 and Task07. Compared with models without pre-training, VoCo pre-training brings average \textcolor{red}{\bm{$+2.98\%$}} DSC improvements.}
\begin{threeparttable}
	\begin{tabular}{c|cccccccccc}
		\toprule[1.2pt]
		\textbf{Method} &\textbf{Task01} &\textbf{Task02} &\textbf{Task03} &\textbf{Task04} &\textbf{Task05} &\textbf{Task06} &\textbf{Task07} &\textbf{Task08} &\textbf{Task09} &\textbf{Task10}\\
        \hline
        nnUNet~\cite{nnunet} &71.25 &91.84 &67.43 &87.08 &79.16 &64.73 &46.08 &68.92 &92.57 &41.69\\
        Swin-B~\cite{swinunetr} &71.74 &92.28 &67.85 &88.66 &72.64 &70.28 &47.88 &64.80 &94.90 &35.13\\
        \hline
        SwinUNETR~\cite{swin} &72.79 &92.06 &64.72 &87.01 &73.76 &71.64 &48.31 &60.72 &95.02 &26.19\\
        SuPrem~\cite{SuPreM} &70.07 &92.55 &68.20 &87.40 &72.92 &72.55 &50.02 &64.71 &96.01 &38.78\\
        \rowcolor{mygray}
        VoCo (nnUNet) &73.56 &\textbf{93.92} &70.09 &\textbf{89.10} &\textbf{80.09} &68.99 &50.43 &\textbf{70.61} &95.70 &41.93\\
        \rowcolor{mygray}
        VoCo (Swin-B) &\textbf{73.94} &93.72 &\textbf{71.22} &88.55 &75.57 &\textbf{75.74} &\textbf{51.34} &67.25 &\textbf{96.12} &\textbf{42.57}\\
        \hline
        \textcolor{cyan}{\bm{$\triangle$}(nnUNet)} &\textcolor{cyan}{$\uparrow$$2.30$}
        &\textcolor{cyan}{$\uparrow$$2.08$}
        &\textcolor{cyan}{$\uparrow$$2.66$}
        &\textcolor{cyan}{$\uparrow$$2.02$}
        &\textcolor{cyan}{$\uparrow$$0.93$}
        &\textcolor{cyan}{$\uparrow$$4.26$}
        &\textcolor{cyan}{$\uparrow$$4.35$}
        &\textcolor{cyan}{$\uparrow$$1.69$}
        &\textcolor{cyan}{$\uparrow$$3.13$}
        &\textcolor{cyan}{$\uparrow$$0.24$}\\
        
        \textcolor{cyan}{\bm{$\triangle$}(Swin-B)} &\textcolor{cyan}{$\uparrow$$2.21$}
        &\textcolor{cyan}{$\uparrow$$2.44$}
        &\textcolor{cyan}{$\uparrow$$3.37$}
        &\textcolor{dark-red}{$\downarrow$$1.11$}
        &\textcolor{cyan}{$\uparrow$$2.93$}
        &\textcolor{cyan}{$\uparrow$$5.46$}
        &\textcolor{cyan}{$\uparrow$$3.46$}
        &\textcolor{cyan}{$\uparrow$$2.45$}
        &\textcolor{cyan}{$\uparrow$$1.22$}
        &\textcolor{cyan}{$\uparrow$$7.44$}\\
        
        \toprule[1.2pt]
	\end{tabular}
    \end{threeparttable}        
\label{table_MSD_performance}
\end{table*}

\begin{table}
	\setlength{\abovecaptionskip}{0.pt}
	\setlength{\belowcaptionskip}{-0.em}
	\centering
	\footnotesize
 \caption{The accuracy (\%) of medical image classification on CC-CCII~\cite{CC-CCII} and LUNA16~\cite{luna} datasets. \underline{Underline} are the baseline performances.}
\begin{threeparttable}
	\begin{tabular}{ccc}
		\toprule[1.2pt]
		\textbf{Method} &\textbf{CC-CCII}~\cite{CC-CCII} &\textbf{LUNA16}~\cite{luna}\\
		\hline
        \textcolor{gray}{\textbf{\emph{From Scratch}}}\\
        3D UNet~\cite{UNET} &89.07 &98.27\\
        \rowcolor{mygray}
        Swin-B~\cite{swinunetr} &\underline{91.04} &\underline{97.54}\\
        \hline
        \textcolor{cyan}{\textbf{\emph{With Pre-training}}}\\
        Jigsaw~\cite{jigsaw} &87.18 &98.07\\
        Rubik++~\cite{rubik2} &89.93 &98.18\\
        PCRLv2~\cite{PCRLv2} &89.35 &98.30\\
        SwinUNETR~\cite{swin} &91.22 &97.41\\
        SuPreM~\cite{SuPreM} &91.83 &97.53\\
        \rowcolor{mygray}
        \textbf{VoCo} &\textbf{93.80}~(\textcolor{cyan}{$\uparrow$$2.76$}) &\textbf{98.67}~(\textcolor{cyan}{$\uparrow$$1.13$})\\
        \toprule[1.2pt]
	\end{tabular}
    \end{threeparttable}        
\label{table_classification}
\end{table}%

\begin{table}
	\setlength{\abovecaptionskip}{0.pt}
	\setlength{\belowcaptionskip}{-0.em}
	\centering
	\footnotesize
 \caption{The DSC (\%) of medical image registration on IXI~\cite{cyclemorph} and OASIS~\cite{marcus2007open} datasets. Some preliminary results are drawn from TransMorph~\cite{Transmorph}. \underline{Underline} are the baseline performances.}
\begin{threeparttable}
	\begin{tabular}{ccc}
		\toprule[1.2pt]
		\textbf{Method} &\textbf{IXI}~\cite{cyclemorph} &\textbf{OASIS}~\cite{marcus2007open}\\
		\hline
        \textcolor{gray}{\textbf{\emph{From Scratch}}}\\
        VoxelMorph~\cite{voxelmorph} &71.5 &78.6\\
        % nnFormer~\cite{nnformer} &72.9 &78.5\\
        Siebert et al~\cite{siebert2021fast} &73.1 &81.0\\
        Mok et al~\cite{mok2021conditional } &73.5 &82.0\\
        TransMorph~\cite{Transmorph} &74.5 &81.6\\
        \rowcolor{mygray}
        Swin-B~\cite{swinunetr} &\underline{72.6} &\underline{81.8}\\
        \hline
        \textcolor{cyan}{\textbf{\emph{With Pre-training}}}\\
        SwinUNETR~\cite{swin} &67.7 &81.5\\
        SuPreM~\cite{SuPreM} &72.9 &81.2\\
        \rowcolor{mygray}
        \textbf{VoCo} &\textbf{73.6}~(\textcolor{cyan}{$\uparrow$$1.0$}) &\textbf{84.4}~(\textcolor{cyan}{$\uparrow$$2.6$})\\
        \toprule[1.2pt]
	\end{tabular}
    \end{threeparttable}        
\label{table_registration}
\end{table}%

\begin{table}
	\setlength{\abovecaptionskip}{0.pt}
	\setlength{\belowcaptionskip}{-0.em}
	\centering
	\footnotesize
 \caption{The performances of the vision-language task: Report Generation on the CTRG dataset~\cite{CTRG}. We use BLEU~\cite{bleu} to measure the accuracy. Some preliminary results only reporting BLEU-4 are drawn from CTRG~\cite{CTRG}.}
\begin{threeparttable}
	\begin{tabular}{ccccc}
		\toprule[1.2pt]
		\textbf{Method} &\textbf{BLEU-1} &\textbf{BLEU-2} &\textbf{BLEU-3} &\textbf{BLEU-4}\\
		\hline
        \textcolor{gray}{\textbf{\emph{From Scratch}}}\\
        % Up-Down~\cite{up-down} &- &- &- &20.7\\
        % AoA~\cite{AoA} &- &- &- &21.2\\
        Mesh-Memor~\cite{Mesh-Memor} &- &- &- &21.0\\
        RSTNet~\cite{rstnet} &- &- &- &18.3\\
        GSKET~\cite{GSKET} &- &- &- &23.5\\
        SL-DG~\cite{CTRG} &- &- &- &23.7\\
        Swin-B~\cite{swinunetr} &58.90 &47.84 &40.71 &35.63\\
        Swin-L~\cite{swinunetr} &59.15 &48.87 &40.89 &36.04\\
        Swin-H~\cite{swinunetr} &59.98 &49.00 &41.28 &36.95\\
        \hline
        \textcolor{cyan}{\textbf{\emph{With Pre-training}}}\\
        TransVW~\cite{TransVW} &54.98 &44.62 &38.34 &33.97\\
        PCRLv2~\cite{PCRLv2} &58.11 &46.96 &40.08 &35.26\\
        SwinUNETR~\cite{swin} &57.15 &46.73 &40.72 &34.28\\
        SuPreM~\cite{SuPreM} &58.70 &48.13 &39.96 &35.23\\
        \rowcolor{mygray}
        \textbf{VoCo}~(Swin-B) &60.45 &49.35 &42.30 &37.42\\
        \rowcolor{mygray}
        \textbf{VoCo}~(Swin-L) &61.79 &\textbf{49.89} &42.25 &36.85\\
        \rowcolor{mygray}
        \textbf{VoCo}~(Swin-H) &\textbf{61.88} &49.82  &\textbf{42.60} &\textbf{37.91}\\
        \toprule[1.2pt]
	\end{tabular}
    \end{threeparttable}        
\label{table_report_generation}
\end{table}%

\begin{table}
	\setlength{\abovecaptionskip}{0.pt}
	\setlength{\belowcaptionskip}{-0.em}
	\centering
	\footnotesize
 \caption{Performances of vocabulary classification and image-text retrieval on CT-RATE~\cite{CT-CLIP} dataset. $\dagger$: \textbf{Note that} CT-CLIP~\cite{CT-CLIP} is based on image-text pre-training.}
\begin{threeparttable}
	\begin{tabular}{ccc}
		\toprule[1.2pt]
		\textbf{Method} &\textbf{Voca. Classi.} (AUC\%) &\textbf{Retrie.} (Recall\%)\\
		\hline
        \textcolor{gray}{\textbf{\emph{From Scratch}}}\\
        3D UNet~\cite{UNET} &53.67 &14.69\\
        Swin-B~\cite{swinunetr} &67.86 &18.12\\
        Swin-L~\cite{swinunetr} &70.89 &21.67\\
        Swin-H~\cite{swinunetr} &70.33 &20.28\\
        \hline
        \textcolor{cyan}{\textbf{\emph{With Pre-training}}}\\
        CT-CLIP$\dagger$~\cite{CT-CLIP} &\textbf{74.70} &23.46\\
        SwinUNETR~\cite{swin} &60.23 &12.51\\
        SuPreM~\cite{SuPreM} &65.18 &20.23\\
        \rowcolor{mygray}
        \textbf{VoCo}~(Swin-B) &71.28 &23.57\\
        \rowcolor{mygray}
        \textbf{VoCo}~(Swin-L) &72.61 &23.79\\
        \rowcolor{mygray}
        \textbf{VoCo}~(Swin-H) &73.69 &\textbf{24.12}\\
        \toprule[1.2pt]
	\end{tabular}
    \end{threeparttable}        
\label{table_vlp}
\end{table}%

\subsection{Comparison with State-of-the-Art Methods}

% \footnote{Checkpoints of pre-trained models are collected by \href{https://github.com/MrGiovanni/SuPreM}{SuPreM}.}

We perform in-depth comparisons with previous methods~\cite{nnunet,Modelgen,dodnet,unimiss,UniMiSS+,swin,clipdriven,SuPreM,rubik2,PCRLv2} that \textbf{have released their codes and checkpoints}. \textbf{Note that} in instances where certain datasets necessitate extensive computational resources or involve limited cases, we exclusively report the results of methods~\cite{nnunet,swin,SuPreM} that with better performances. Our evaluations span across segmentation, classification, registration, and vision-language tasks. In following discussions, the term \textbf{\emph{baseline} denotes adopting the same backbones but without pre-training (from scratch)}.

\subsubsection{Medical Image Segmentation}

\indent \textbf{Seven widely-used segmentation datasets}. As shown in Table~\ref{table_abdomen_performance}, on seven widely-used segmentation datasets, VoCo demonstrates leading performances and surpass previous methods~\cite{nnunet,Modelgen,dodnet,unimiss,UniMiSS+,swin,clipdriven,SuPreM} by a clear margin. It can be seen that the general method MoCo-v3~\cite{moco,mocov3} did not perform well on medical tasks. Since MoCo v3~\cite{moco,mocov3} heavily relies on a large batch size to acquire adequate negative samples, which is not feasible in 3D medical images. Moreover, the negative relation between different images used in MoCo v3~\cite{moco,mocov3} is not appropriate in medical images.
 
 Notably, VoCo outperforms the baseline by average \bm{$3.12\%$} DSC. SuPreM~\cite{SuPreM} achieved the best results among the previous pre-training methods since it used an abdomen dataset~\cite{atlas} for supervised pre-training and the datasets in Table~\ref{table_abdomen_performance} are almost abdomen datasets. VoCo surpasses SuPreM~\cite{SuPreM} and achieves new state-of-the-art performances. Specifically, for the challenging ToTalSegmentor~\cite{Total} dataset, VoCo (Swin-H) outperforms SuPreM~\cite{SuPreM} by \bm{$3.22\%$} DSC. The overall results in Table~\ref{table_abdomen_performance} vividly underscore the effectiveness of our method.

\textbf{24 organ/tumor segmentation tasks}. As shown in Table~\ref{table_other_downstream_performance}, we report the results on 24 organ and tumor segmentation datasets, across different modalities and anatomical regions as shown in Table~\ref{table_downstream_dataset}. Notably, models with VoCo pre-training outperform those without pre-training by average \bm{$4.42\%$} DSC. It is worth noting that a majority of these datasets contain fewer than 50 annotated cases for fine-tuning, highlighting the effectiveness of pre-training as a label-efficient solution. The overall improvements observed across these 24 datasets serve as compelling evidence for the efficacy of our proposed large-scale pre-training method.

\textbf{MSD Challenge}. Table~\ref{table_MSD_performance} reports the results on the MSD 10-Task~\cite{MSD} dataset. We adopt the settings of nnUNet~\cite{nnunet} for fair comparisons. Notably, with VoCo pre-training, the segmentation DSC is improved by average \bm{$2.98\%$}.

\subsubsection{Medical Image Classification}

The medical image classification results on CC-CCII~\cite{CC-CCII} and LUNA16~\cite{luna} are shown in Table~\ref{table_classification}. Given the near-optimal accuracy of lung nodule detection on LUNA16~\cite{luna}, the benefits of pre-training are not obvious. For Covid classification on CC-CCII~\cite{CC-CCII}, VoCo outperforms the baseline by \bm{$2.76\%$} and SuPreM~\cite{SuPreM} by \bm{$1.97\%$}. Notably, SuPrem~\cite{SuPreM} conducted supervised segmentation pre-training on only abdomen datasets, potentially limiting its transferability to chest classification tasks.

\subsubsection{Medical Image Registration}

The medical image registration results on IXI~\cite{cyclemorph} and OASIS~\cite{marcus2007open} datasets are shown in Table~\ref{table_registration}. We adopt TransMorph~\cite{Transmorph} as the baseline. \textbf{Note that} in this paper we focus on evaluating the effectiveness of pre-training, thus we did not propose new registration algorithms. Thus, our registration analyses emphasize backbone comparisons (scratch versus pre-trained). We find that previous pre-training methods~\cite{swin,SuPreM} did not perform well on registration. While on brain MRI registration dataset OASIS~\cite{marcus2007open}, VoCo brings \bm{$2.6\%$} DSC improvements, which is a non-trivial improvement in registration.

\subsubsection{Vision-Language Analysis}

As shown in Table~\ref{table_downstream_compare}, this study pioneers the assessment of medical image pre-training efficacy in Vision-Language (VL) tasks. Specifically, we evaluate the report generation task on CTRG-Chest~\cite{m2kt} and extend the evaluation to vocabulary classification and report-volume retrieval on the CT-RATE~\cite{CT-CLIP} dataset. The results are shown in Tables~\ref{table_report_generation} and~\ref{table_vlp}. \textbf{Note that} in this paper we focus on medical image pre-training, thus we verify the effectiveness via replacing the vision encoders. For the language models, we maintain the original settings from M2KT~\cite{m2kt} and CT-CLIP~\cite{CT-CLIP} for CTRG-Chest and CT-RATE, respectively.

VoCo attains superior performances compared to previous medical image pre-training methods~\cite{SuPreM,swin,TransVW,PCRLv2}. Specifically, for report generation in Table~\ref{table_report_generation}, VoCo (Swin-H) achieves the highest score BLEU-4 (37.91\%). In Table~\ref{table_vlp}, VoCo (Swin-H) achieves 73.69\% AUC in vocabulary classification and 24.12\% recall in report-volume retrieval. Although SuPreM~\cite{SuPreM} performs well in abdomen segmentation datasets, it falls short in enhancing chest VL tasks. The results from VoCo underscore the significance of a robust vision encoder in VL tasks, which can provide more precise visual information for language models.

\subsubsection{Discussion}~\label{sec_discussion}

\begin{figure}
	\centering
	\includegraphics[width=1\linewidth]{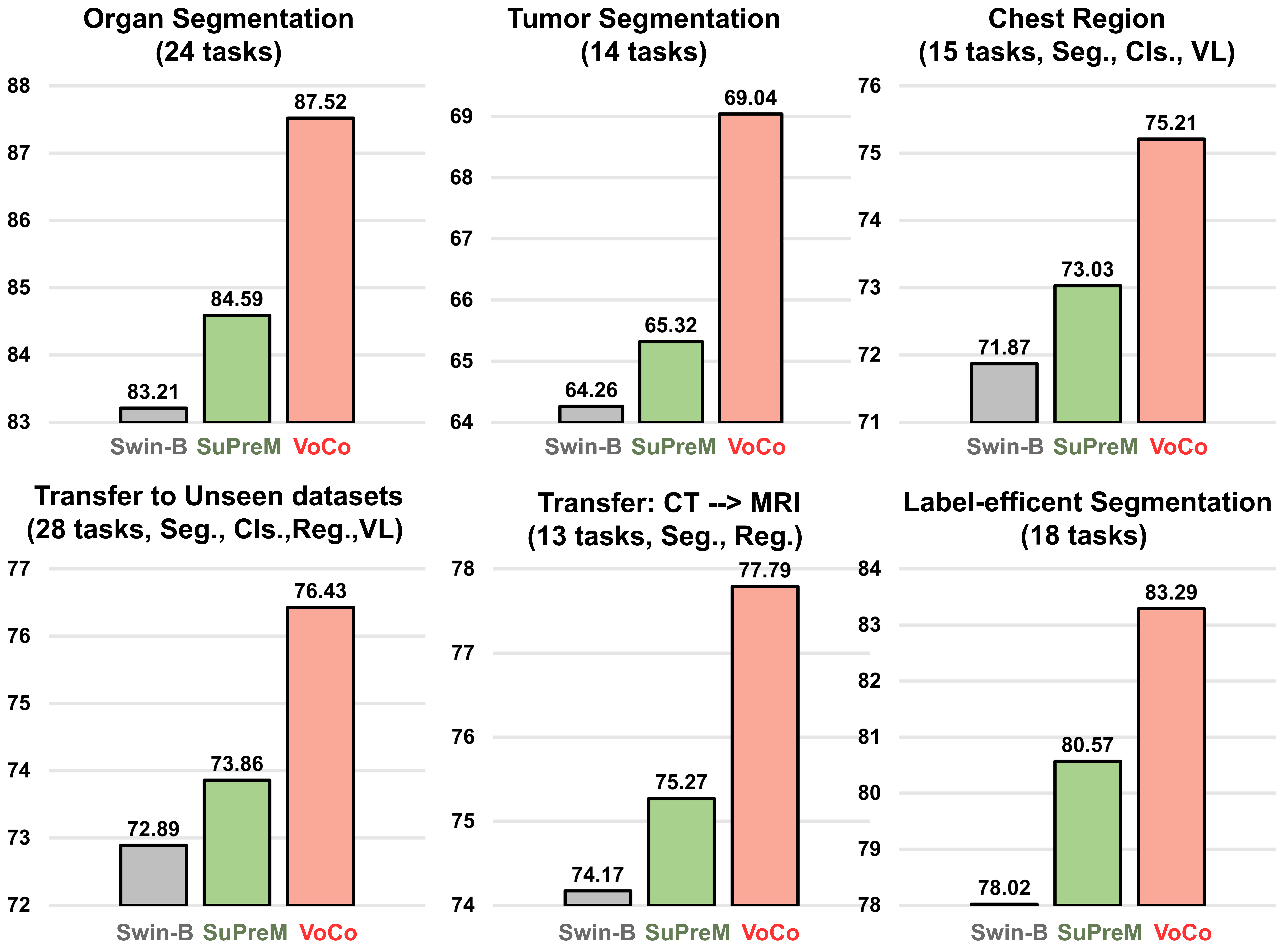}
	\caption{\textbf{Overall comparisons}. Swin-B denotes using the randomly initialized SwinUNETR~\cite{swinunetr} as the backbone. Both SuPreM~\cite{SuPreM} and VoCo use Swin-B~\cite{swinunetr} as backbones for pre-training. Given the significant representation of chest datasets within our benchmark, we present the enhancement outcomes across 15 chest analysis tasks.}
	\label{fig_improve}
\end{figure}

\textbf{Overall improvements}. As shown in Fig.~\ref{fig_improve}, with the same backbone, VoCo outperforms the baseline (from scratch) by a clear margin. SuPreM~\cite{SuPreM} emerged as the top performer among previous pre-training methods~\cite{PCRLv2,clipdriven,Modelgen,unimiss,UniMiSS+,swin,dodnet,nnunet,rubik2,TransVW}. Specifically, VoCo surpasses SuPreM~\cite{SuPreM} by an average of \textbf{2.93\%}, \textbf{3.72\%}, \textbf{2.57\%}, \textbf{2.18\%}, \textbf{3.52\%}, and \textbf{2.72\%} on 24 organ segmentation datasets, 14 tumor segmentation datasets, 15 chest analysis datasets, 28 unseen datasets, 13 cross-modal datasets, and 18 label-efficient segmentation datasets, respectively. Consistent improvements on 48 downstream datasets provide strong evidence of the effectiveness of VoCo. 

\begin{figure}
	\centering
	\includegraphics[width=1\linewidth]{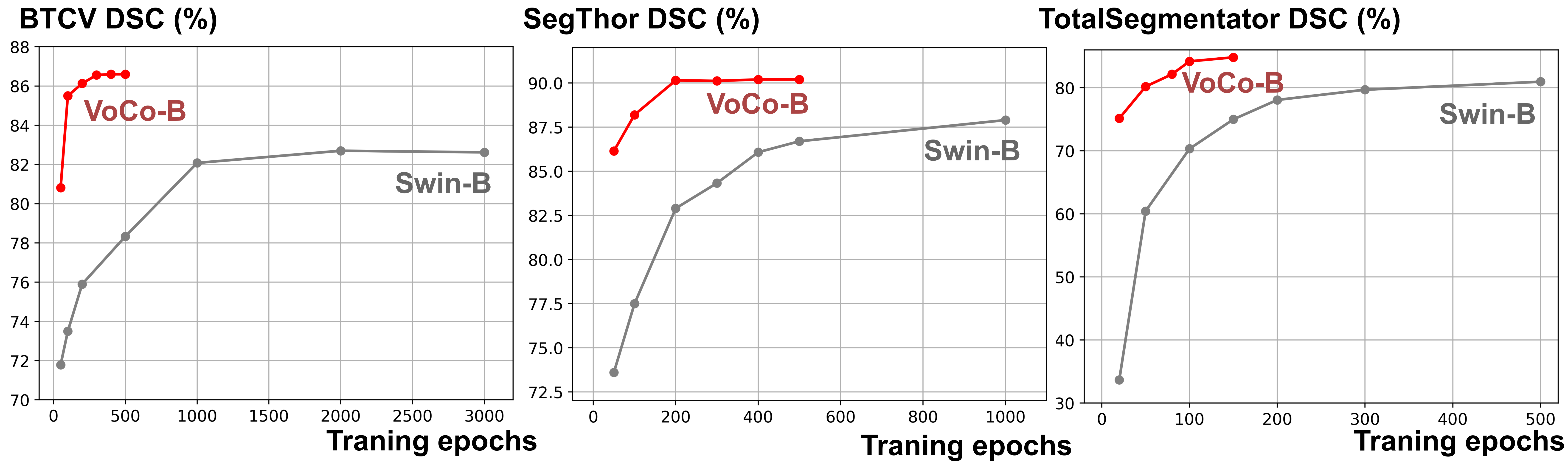}
	\caption{\textbf{Efficient finetuning}. Analysis on BTCV~\cite{btcv}, SegThor~\cite{segthor}, and TotalSegmentator~\cite{Total}, where SegThor~\cite{segthor} is unseen in pre-training. Compared with the randomly initialized backbone Swin-B~\cite{swinunetr}, VoCo achieves higher accuracy within fewer training epochs.}
	\label{fig_convergence}
\end{figure}

\textbf{Transferability to unseen datasets}. As shown in Table~\ref{table_downstream_dataset}, our evaluation benchmark encompasses 28 datasets unseen in pre-training. As shown in Fig.~\ref{fig_improve}, VoCo demonstrates an average improvement of \textbf{3.53\%} over the baseline Swin-B~\cite{swinunetr} when evaluated across these 28 unseen datasets.

\textbf{Transferability to unseen modality}. We conduct pre-training on CT datasets and subsequently transfer the learned models to another 3D medical imaging modality, \emph{i.e.}, MRI. As shown in Table~\ref{table_downstream_dataset}, our benchmark encompasses 13 MRI datasets spanning various tasks such as segmentation and registration. As shown in Fig.~\ref{fig_improve}, VoCo yields an average improvement of \textbf{3.52\%} across these 13 datasets, underscoring its efficacy in facilitating cross-modal transferability.

\textbf{Label-efficient solution}. In 3D medical image analysis, many datasets suffer from the scarcity of labeled data, primarily due to the substantial costs of annotation. As shown in Table~\ref{table_downstream_dataset}, there are 18 segmentation datasets with less than 50 labeled cases for finetuning. As shown in Fig.~\ref{fig_improve}, VoCo emerges as a label-efficient solution tailored for datasets constrained by limited labeled data, consistently delivering superior performances.

\textbf{Pre-trained backbones}. We use both nnUNet~\cite{nnunet} and SwinUNETR~\cite{swinunetr} for pre-training. Although nnUNet~\cite{nnunet} emerged as a strong segmentation baseline, it is not a scalable network architecture~\cite{stunet}, with only 31M model params. Thus, we primarily focus on investigating the scaling law of SwinUNETR. Our analysis reveals that the pre-trained SwinUNETR~\cite{swinunetr} exhibits more substantial enhancements compared to the pre-trained nnUNet, \emph{i.e.}, \textbf{+3.34\%} and \textbf{+1.98\%} DSC on 34 segmentation datasets (Table~\ref{table_other_downstream_performance} and \ref{table_MSD_performance}). The relatively modest improvements observed in pre-trained nnUNet could potentially stem from variations in pre-processing strategies, given nnUNet's reliance on a distinct data-fingerprint processing technique.

\textbf{Efficient finetuning}. Previous works~\cite{dino,dinov2,SuPreM} proved that strong pre-training models can notably expedite training convergence, resulting in improved performance with fewer training epochs. As shown in Fig.~\ref{fig_convergence}, VoCo substantially expedites the training convergence speed on BTCV~\cite{btcv}, Segthor~\cite{segthor}, and TotalSegmentator~\cite{Total}, and this phenomenon is generalized in all 48 tasks. This is a non-trivial contribution to efficient finetuning, particularly beneficial for datasets demanding extensive computational resources~\cite{Total}. Our pre-trained models are poised to save computation costs in medical image analysis, making a strong step towards efficient learning.

\textbf{Failure cases}. Although consistent improvements (at least 1\%) are observed on 48 datasets, marginal improvements persist in a handful of cases. Specifically, VoCo gains less than 1.5\% improvements on 5 of 48 datasets. The presence of challenging datasets, \emph{e.g.}, Positron Emission Tomography (PET) dataset AutoPET~\cite{AutoPET} poses unique obstacles, primarily due to their distinct imaging characteristics compared to our pre-training datasets. These differences result in domain gaps that constrain the effectiveness of our pre-training.

\begin{figure*}
	\centering
	\includegraphics[width=0.85\linewidth]{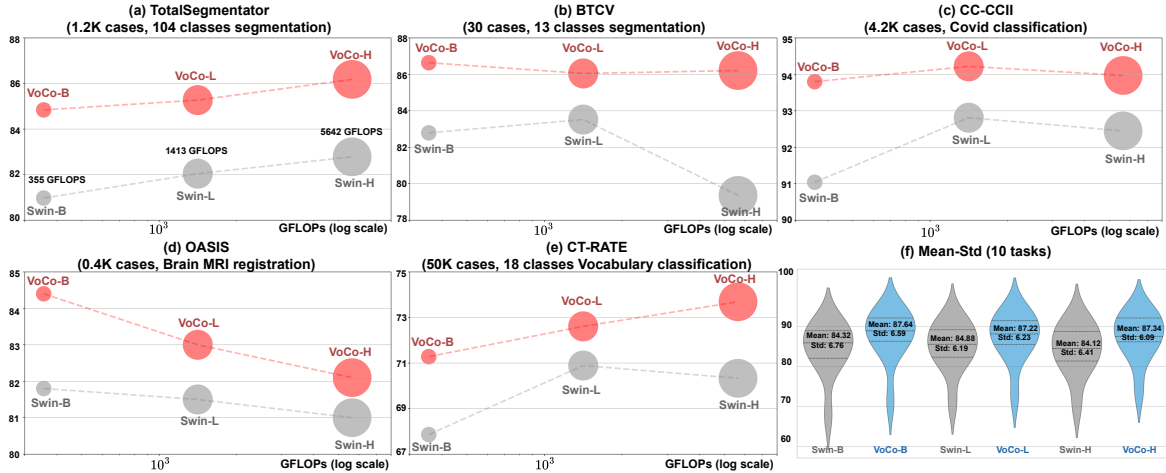}
	\caption{\emph{\textbf{Are larger models always better?}} The answer appears to be \textbf{\emph{no}}. We present the scaling results of TotalSegmentator~\cite{Total}, BTCV~\cite{btcv}, CC-CCII~\cite{CC-CCII}, OASIS~\cite{marcus2007open}, and CT-RATE~\cite{CT-CLIP} in (a)-(e), respectively, covering various downstream tasks. We compared our pre-trainined models with the randomly initialized models~\cite{swinunetr}, taking into account both accuracy and computation costs (GFLOPs computed for a $96{\times}96{\times}96$ size of volume, shown in (a)). (f) presents the mean and standard deviation (STD) values across 10 downstream tasks~\cite{btcv,amos,FLARE22,word,Total,atlas,CC-CCII,CT-CLIP,marcus2007open}.
 }
	\label{fig_scaling}
\end{figure*}

\subsection{Scaling Law in Medical Image Analysis}

\textbf{Are larger models always better?} In medical tasks, the answer appears to be \textbf{\emph{no}}. It can be observed from Fig.~\ref{fig_scaling} that for some specific tasks, models with smaller sizes can achieve better performances. In this paper, we delves into factors affecting the model capacity scaling law, including: \emph{number of finetuning cases, data diversity, and task difficulties}.

As shown in Fig.~\ref{fig_scaling}, \textbf{(a)} TotalSegmentator~\cite{Total} is a challenging dataset,  containing 1.2K cases and 104 classes for segmentation. In this case, the largest model VoCo-H yields the best results. \textbf{(b)} BTCV~\cite{btcv} is with only 24 cases for finetuning, potentially leading larger models to \textbf{overfit} on limited data, thus hindering validation performance. \textbf{(c)} Although CC-CCII~\cite{CC-CCII} encompasses 4.2K cases for training, it is a simple binary classification task (over 90\% accuracy), suggesting that excessively large models may not be necessary. \textbf{(d)} OASIS~\cite{marcus2007open} is brain MRI datasets with only 0.4K cases for registration and it also lacks significant structural diversity. In this case, the smallest VoCo-B delivers the best results. \textbf{(e)} CT-Rate~\cite{CT-CLIP} is with 50K cases for 18 classes vocabulary classification. Given large-scale data for training, larger models demonstrate higher performances.

\begin{table}
	\setlength{\abovecaptionskip}{0.pt}
	\setlength{\belowcaptionskip}{-0.em}
	\centering
	\footnotesize
\begin{threeparttable}
	\begin{tabular}{ccc|ccccc}
		\toprule[1.2pt]
		\multicolumn{2}{c}{\textbf{Self}} &\multirow{2}{*}{\textbf{Semi}} &\multirow{2}{*}{\textbf{Total}} &\multirow{2}{*}{\textbf{BTCV}} &\multirow{2}{*}{\textbf{CCII}} &\multirow{2}{*}{\textbf{OAS.}} &\multirow{2}{*}{\textbf{CTRG}}\\
        \cline{1-2}
        \textbf{intra} &\textbf{inter} &\\
		\hline
  \XSolidBrush &\XSolidBrush &\XSolidBrush &80.97 &82.79 &91.04 &81.79 &58.90\\
        \CheckmarkBold &\XSolidBrush &\XSolidBrush &81.38 &84.51 &92.85 &82.34 &59.37\\
        \CheckmarkBold &\CheckmarkBold &\XSolidBrush &82.07 &85.42 &93.64 &82.49 &60.23\\
        \rowcolor{mygray}
        \XSolidBrush &\XSolidBrush &\CheckmarkBold &84.02 &85.37 &91.98 &82.12 &59.13\\
        \rowcolor{pink}
        \CheckmarkBold &\CheckmarkBold &\CheckmarkBold &\textbf{84.84} &\textbf{86.64} &\textbf{93.80} &\textbf{84.43} &\textbf{60.45}\\
        \toprule[1.2pt]
	\end{tabular}
    \end{threeparttable}        
	\caption{Evaluation of self- and semi-supervised learning. We report the downstream results of VoCo-B on Total.~\cite{Total}, BTCV~\cite{btcv}, CCII~\cite{CC-CCII}, OASIS~\cite{marcus2007open}, and CTRG~\cite{m2kt}.}
\label{table_ablation}
\end{table}%

\begin{figure}
	\centering
	\includegraphics[width=0.75\linewidth]{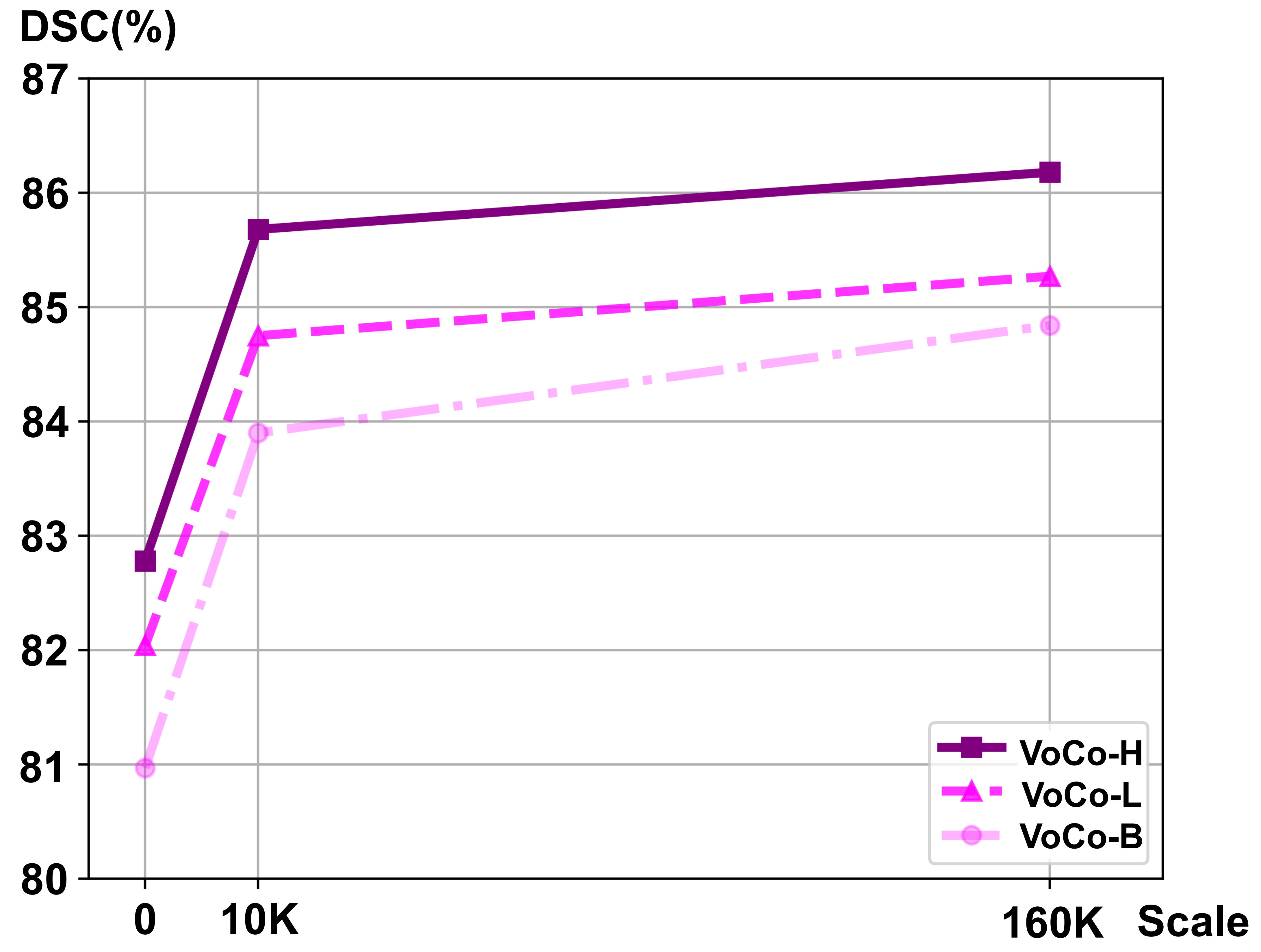}
	\caption{\textbf{Data scaling law}. We scale up the data from 10K to 160K and report the DSC (\%) of TotalSegmentator~\cite{Total}.}
	\label{fig_data_scale}
\end{figure}

\textbf{Tailor different model sizes to various medical
tasks}. Drawing from experimental insights discussed above, we empirically propose simple and reasonable guidelines for tailoring various medical tasks: \textbf{(1)} Tasks with extensive labeled data for fine-tuning potentially benefit from larger models. \textbf{(2)} Tasks spanning diverse anatomical regions potentially benefit from larger models. \textbf{(3)} Tasks requiring recognition across a higher number of classes (more challenging) are better addressed with larger models.

Although these guidelines have been assessed on our comprehensive benchmark, they may not universally apply to all medical tasks given the substantial diversity within the medical domain. Thus, we release pre-trained models of varying sizes to aid researchers in selecting the most appropriate models for their specific needs.

\subsection{Ablation Studies}\label{sec_ablation}

Our preliminary investigation VoCo-v1~\cite{VoCo} has provided fundamental ablation studies, focusing on exploring various loss functions and hyperparameter configurations. Compared with VoCo-v1~\cite{VoCo}, we further evaluate the effectiveness of volume contrast, omni-supervised learning, and data scaling from 10K to 160K. 
We use Swin-B~\cite{swinunetr} as the backbone and present the results on diverse datasets, including TotalSegmentator~\cite{Total}, BTCV~\cite{btcv}, CC-CCII~\cite{CC-CCII}, OASIS~\cite{marcus2007open}, and CTRG~\cite{m2kt}, across segmentation, classification, registration, and vision-language tasks.

\textbf{Volume contrast}. As shown in Table~\ref{table_ablation}, inter-volume contrast consistently enhances performance across five datasets. The combination of intra- and inter-volume contrast can yield higher improvements compared with the randomly initialized backbone~\cite{swinunetr}.

\textbf{Omni-supervised pre-training}. As shown in Table~\ref{table_ablation}, semi-supervised learning can effectively improve the performances. Specifically, for TotalSegmentator~\cite{Total}, it leads to substantial DSC improvements from \bm{$82.07\%$} to \bm{$84.84\%$}, which is a non-trivial boost in this challenging segmentation dataset. It is worth noting that the pure semi-supervised pre-training can achieve competitive results on segmentation tasks~\cite{Total,btcv}, but it does not improve significantly in classification, registration, and VL tasks. Combined with self- and semi-supervised learning, the omni-supervised pre-training can achieve the best performances.

\textbf{Data scaling law in medical image pre-training}. We scale up the pre-training data (Table~\ref{table_dataset}) from 10K to 160K and present the findings on the TotalSegmentator~\cite{Total} dataset in Fig.~\ref{fig_data_scale}, showcasing the impact of expanding the pre-training dataset. Notably, the enhancements from 10K to 160K appear marginal. This phenomenon could be attributed to factors like data quality and diversity, network scalability, or nearing the upper limit of improvement.

\begin{figure*}
	\centering
	\includegraphics[width=0.95\linewidth]{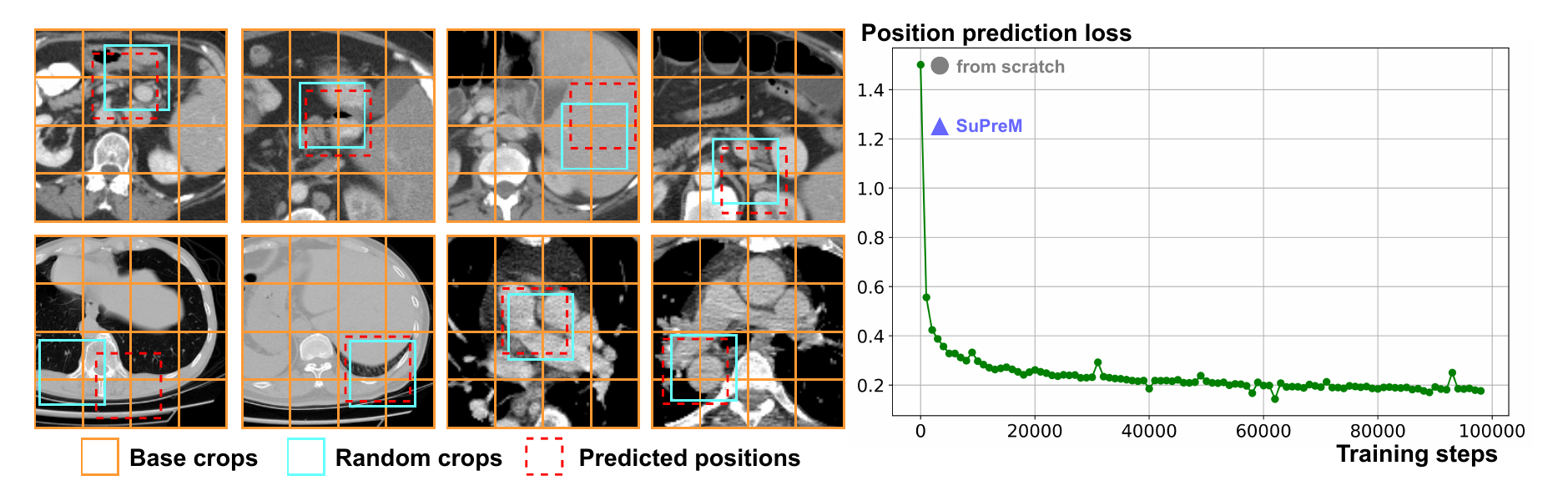}
	\caption{\textbf{Case study for contextual position prediction}. \textbf{(1)} The left part shows the contextual position prediction results. Specifically, we set thresholds for the prediction logits to output the most probable positions. The predictions closely match the original positions of random crops. The bottom left is a failure case where two regions share similar structures. \textbf{(2)} As shown in the right part, the position prediction loss converges steadily during pre-training. We further verify the position prediction results of the model from scratch and the pre-trained SuPreM~\cite{SuPreM} model. Notably, through supervised segmentation pre-training, SuPreM~\cite{SuPreM} also enhances the contextual position prediction capability, implicitly indicating a positive correlation between segmentation performance and our proposed contextual position prediction.
 }
	\label{fig_case_study}
\end{figure*}

\begin{figure*}
	\centering
	\includegraphics[width=0.95\linewidth]{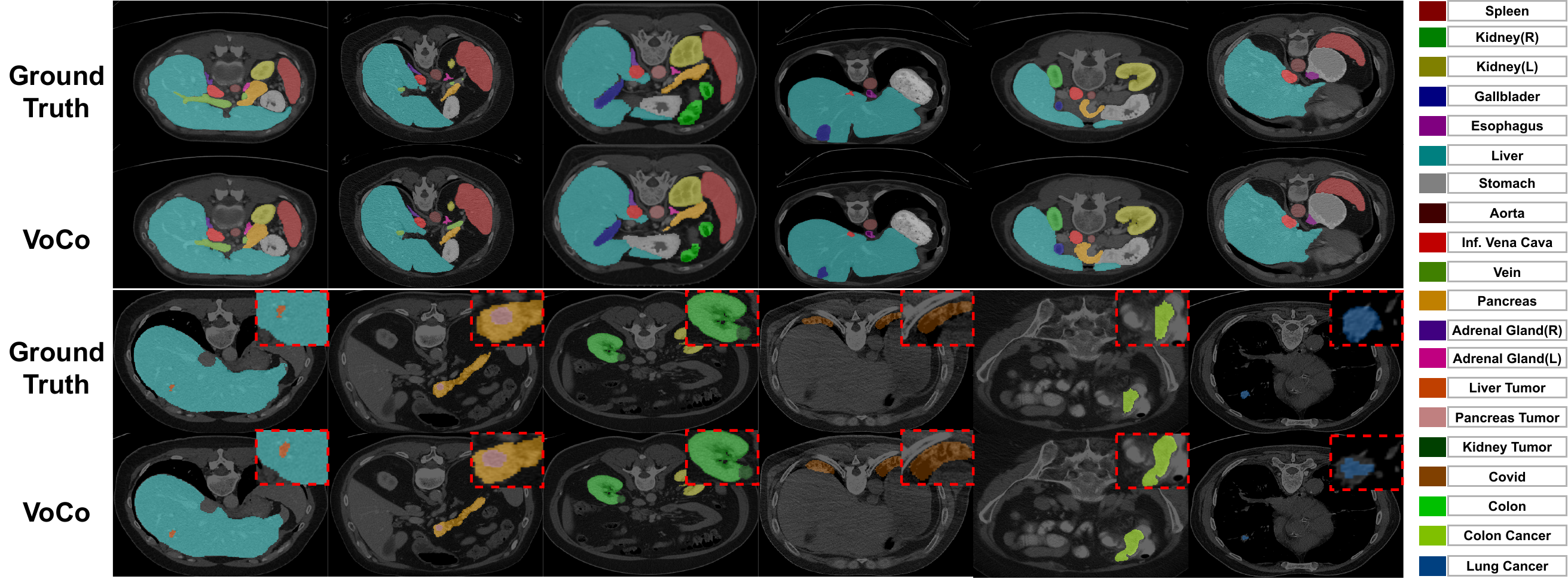}
	\caption{Qualitative segmentation results on~\cite{FLARE22,btcv,MSD,kits,covid}. Tumor regions are zoomed in for better visualization. 
 }
	\label{fig_vis}
\end{figure*}

\subsection{Qualitative Visualization Results}

\textbf{Contextual Position Prediction}. As shown in Fig.~\ref{fig_case_study}, we present some visualization results of contextual position prediction. The loss for contextual position prediction converges steadily during pre-training. The position predictions generated by VoCo closely align with the ground truth, underscoring the efficacy of our proposed pretext task.

\textbf{Qualitative segmentation results}. We present some visualization results in Fig.~\ref{fig_vis}, which covers different anatomical regions. The visualization results demonstrate that our method can broadly apply to various downstream tasks.

\section{Limitations and Future Directions}
\label{sec_limitation}

Although our pre-training method has demonstrated promising results across various medical tasks, there are still several limitations that can be further explored in the future:
\begin{itemize}

\item \textbf{Data engines for improving data quality}. The improvements become marginal when scaling the data from 10K to 160K. Although we have curated and pre-processed the pre-training dataset, the PreCT-160K still inevitably includes numerous low-quality cases.
Data quality plays a pivotal role in pre-training to fully leverage the potential of large-scale datasets~\cite{sam,depthanything,dinov2,freemask}. In the future, we will focus on constructing data engines to improve the quality of datasets.

\item \textbf{Data diversity to encompass distinctive image characteristics}. As discussed in Sec.~\ref{sec_discussion}, VoCo shows marginal enhancements in a few downstream tasks characterized by unique imaging features. Given the extensive diversity of medical datasets, we will further enhance the diversity of our pre-training dataset in future endeavors.

\item \textbf{Multi-modal pre-training}. In this study, we exclusively utilize CT data for 3D medical image pre-training. In the future, we will also build a large-scale MRI pre-training dataset and combine with CT to facilitate multi-modal 3D medical image pre-training. 

\item \textbf{Scalable network architectures}. This study does not center on crafting novel network architectures for 3D medical image analysis. The current backbones~\cite{nnunet,swinunetr} we employ may not exhibit scalability for large-scale pre-training. In the future, we will delve into the development of advanced network architectures for scalable pre-training.

\item \textbf{Advance omni-supervised learning strategies}. As in Sec.~\ref{sec_ablation}, the omni-supervised pre-training yields better performances than the pure SSL method. In the future, our focus will shift toward omni-supervised learning techniques to effectively leverage both labeled and unlabeled data.

\end{itemize}

\section{Conclusion}

In this paper, we proposed a simple-yet-effective \textbf{Vo}lume \textbf{Co}ntrast (\textbf{VoCo}) framework for large-scale 3D medical image pre-training. Inspired by the consistent geometric relation between different organs, we propose to leverage the geometric context priors to learn consistent semantic representations for SSL. VoCo can also be seamlessly integrated into a semi-supervised learning framework for omni-supervised pre-training. To facilitate the study of large-scale 3D medical image pre-training, we curated the existing largest medical image pre-training dataset PreCT-160K, which encompasses 160K CT volumes (42M slices) covering diverse anatomical structures. We further delve into the scaling law of model capacity and propose the guidelines for tailoring different model sizes to various medical tasks. To evaluate the effectiveness of pre-training, we establish a comprehensive evaluation benchmark encompassing 48 downstream datasets across various tasks. Extensive experiments highlighted the superior performances of VoCo compared with previous methods.

\ifCLASSOPTIONcompsoc
  \section*{Acknowledgments}
\else
  \section*{Acknowledgment}
\fi
\noindent This work was supported by Hong Kong Innovation and Technology Fund (Project No. ITS/028/21FP and No. MHP/002/22), and Research Grants Council of the Hong Kong Special Administrative Region, China (Project No. T45-401/22-N).

% \rebut{We highly appreciate the editors and reviewers for their valuable comments on our manuscript, which significantly improve our manuscript.}

\bibliography{ref}

\begin{thebibliography}{100}
\providecommand{\url}[1]{#1}
\csname url@samestyle\endcsname
\providecommand{\newblock}{\relax}
\providecommand{\bibinfo}[2]{#2}
\providecommand{\BIBentrySTDinterwordspacing}{\spaceskip=0pt\relax}
\providecommand{\BIBentryALTinterwordstretchfactor}{4}
\providecommand{\BIBentryALTinterwordspacing}{\spaceskip=\fontdimen2\font plus
\BIBentryALTinterwordstretchfactor\fontdimen3\font minus \fontdimen4\font\relax}
\providecommand{\BIBforeignlanguage}[2]{{%
\expandafter\ifx\csname l@#1\endcsname\relax
\typeout{** WARNING: IEEEtran.bst: No hyphenation pattern has been}%
\typeout{** loaded for the language `#1'. Using the pattern for}%
\typeout{** the default language instead.}%
\else
\language=\csname l@#1\endcsname
\fi
#2}}
\providecommand{\BIBdecl}{\relax}
\BIBdecl

\bibitem{VoCo}
L.~Wu, J.~Zhuang, and H.~Chen, ``Voco: A simple-yet-effective volume contrastive learning framework for 3d medical image analysis,'' in \emph{CVPR}, 2024.

\bibitem{abdomenct1k}
J.~Ma \emph{et~al.}, ``Abdomenct-1k: Is abdominal organ segmentation a solved problem?'' \emph{TPAMI}, vol.~44, no.~10, pp. 6695--6714, 2021.

\bibitem{qiu2023rethinking}
Z.~Qiu \emph{et~al.}, ``Rethinking dual-stream super-resolution semantic learning in medical image segmentation,'' \emph{TPAMI}, 2023.

\bibitem{medley2021cycoseg}
D.~O. Medley \emph{et~al.}, ``Cycoseg: A cyclic collaborative framework for automated medical image segmentation,'' \emph{TPAMI}, vol.~44, no.~11, pp. 8167--8182, 2021.

\bibitem{azad2024medical}
R.~Azad \emph{et~al.}, ``Medical image segmentation review: The success of u-net,'' \emph{TPAMI}, 2024.

\bibitem{xie2023learning}
Y.~Xie \emph{et~al.}, ``Learning from partially labeled data for multi-organ and tumor segmentation,'' \emph{TPAMI}, 2023.

\bibitem{wu2022minimizing}
F.~Wu and X.~Zhuang, ``Minimizing estimated risks on unlabeled data: A new formulation for semi-supervised medical image segmentation,'' \emph{TPAMI}, vol.~45, no.~5, pp. 6021--6036, 2022.

\bibitem{PCRLv2}
H.-Y. Zhou \emph{et~al.}, ``A unified visual information preservation framework for self-supervised pre-training in medical image analysis,'' \emph{TPAMI}, 2023.

\bibitem{UniMiSS+}
Y.~Xie \emph{et~al.}, ``Unimiss+: Universal medical self-supervised learning from cross-dimensional unpaired data,'' \emph{TPAMI}, 2024.

\bibitem{SSL3D}
A.~Taleb \emph{et~al.}, ``3d self-supervised methods for medical imaging,'' \emph{NeurIPS}, vol.~33, pp. 18\,158--18\,172, 2020.

\bibitem{annotating}
K.~Gr{\"u}nberg \emph{et~al.}, ``Annotating medical image data,'' \emph{MedIA}, pp. 45--67, 2017.

\bibitem{dino}
M.~Caron \emph{et~al.}, ``Emerging properties in self-supervised vision transformers,'' in \emph{ICCV}, 2021.

\bibitem{ibot}
J.~Zhou \emph{et~al.}, ``ibot: Image bert pre-training with online tokenizer,'' \emph{arXiv preprint arXiv:2111.07832}, 2021.

\bibitem{simclr}
T.~Chen \emph{et~al.}, ``A simple framework for contrastive learning of visual representations,'' in \emph{ICML}, 2020, pp. 1597--1607.

\bibitem{mocov3}
X.~Chen \emph{et~al.}, ``An empirical study of training self-supervised vision transformers,'' \emph{arXiv preprint arXiv:2104.02057}, 2021.

\bibitem{mim}
J.~Zhuang \emph{et~al.}, ``Mim: Mask in mask self-supervised pre-training for 3d medical image analysis,'' \emph{arXiv preprint arXiv:2404.15580}, 2024.

\bibitem{swin}
Y.~Tang \emph{et~al.}, ``Self-supervised pre-training of swin transformers for 3d medical image analysis,'' in \emph{CVPR}, 2022, pp. 20\,730--20\,740.

\bibitem{Alice}
Y.~Jiang \emph{et~al.}, ``Anatomical invariance modeling and semantic alignment for self-supervised learning in 3d medical image analysis,'' in \emph{ICCV}, 2023, pp. 15\,859--15\,869.

\bibitem{continual}
Y.~Ye \emph{et~al.}, ``Continual self-supervised learning: Towards universal multi-modal medical data representation learning,'' in \emph{CVPR}, 2024, pp. 11\,114--11\,124.

\bibitem{dinov2}
M.~Oquab \emph{et~al.}, ``Dinov2: Learning robust visual features without supervision,'' \emph{TMLR}, 2023.

\bibitem{MAE}
K.~He \emph{et~al.}, ``Masked autoencoders are scalable vision learners,'' in \emph{CVPR}, 2022, pp. 16\,000--16\,009.

\bibitem{sam}
A.~Kirillov \emph{et~al.}, ``Segment anything,'' in \emph{ICCV}, 2023, pp. 4015--4026.

\bibitem{internvl}
Z.~Chen \emph{et~al.}, ``Internvl: Scaling up vision foundation models and aligning for generic visual-linguistic tasks,'' in \emph{CVPR}, 2024, pp. 24\,185--24\,198.

\bibitem{depthanything}
L.~Yang \emph{et~al.}, ``Depth anything: Unleashing the power of large-scale unlabeled data,'' in \emph{CVPR}, 2024.

\bibitem{PCRLv1}
H.-Y. Zhou \emph{et~al.}, ``Preservational learning improves self-supervised medical image models by reconstructing diverse contexts,'' in \emph{ICCV}, 2021, pp. 3499--3509.

\bibitem{SuPreM}
W.~Li, A.~Yuille, and Z.~Zhou, ``How well do supervised models transfer to 3d image segmentation?'' in \emph{ICLR}, 2024.

\bibitem{unimiss}
Y.~Xie \emph{et~al.}, ``Unimiss: Universal medical self-supervised learning via breaking dimensionality barrier,'' in \emph{ECCV}, 2022, pp. 558--575.

\bibitem{geo}
Y.~He \emph{et~al.}, ``Geometric visual similarity learning in 3d medical image self-supervised pre-training,'' in \emph{CVPR}, 2023, pp. 9538--9547.

\bibitem{atlas}
C.~Qu \emph{et~al.}, ``Abdomenatlas-8k: Annotating 8,000 ct volumes for multi-organ segmentation in three weeks,'' \emph{NeurIPS}, vol.~36, 2024.

\bibitem{rubik}
X.~Zhuang \emph{et~al.}, ``Self-supervised feature learning for 3d medical images by playing a rubik’s cube,'' in \emph{MICCAI}, 2019, pp. 420--428.

\bibitem{MAE3D}
Z.~Chen \emph{et~al.}, ``Masked image modeling advances 3d medical image analysis,'' in \emph{WACV}, 2023, pp. 1970--1980.

\bibitem{depthv2}
L.~Yang \emph{et~al.}, ``Depth anything v2,'' \emph{NeurIPS}, 2024.

\bibitem{imagenet}
J.~Deng \emph{et~al.}, ``Imagenet: A large-scale hierarchical image database,'' in \emph{CVPR}, 2009, pp. 248--255.

\bibitem{swav}
M.~o. Caron, ``Unsupervised learning of visual features by contrasting cluster assignments,'' \emph{NeurIPS}, vol.~33, pp. 9912--9924, 2020.

\bibitem{larsson2017colorization}
G.~Larsson \emph{et~al.}, ``Colorization as a proxy task for visual understanding,'' in \emph{CVPR}, 2017, pp. 6874--6883.

\bibitem{pathak2016context}
D.~Pathak \emph{et~al.}, ``Context encoders: Feature learning by inpainting,'' in \emph{CVPR}, 2016, pp. 2536--2544.

\bibitem{chen2020generative}
M.~Chen \emph{et~al.}, ``Generative pretraining from pixels,'' in \emph{ICML}, 2020, pp. 1691--1703.

\bibitem{moco}
K.~He \emph{et~al.}, ``Momentum contrast for unsupervised visual representation learning,'' in \emph{CVPR}, 2020, pp. 9729--9738.

\bibitem{byol}
J.-B. Grill \emph{et~al.}, ``Bootstrap your own latent-a new approach to self-supervised learning,'' \emph{NeurIPS}, vol.~33, pp. 21\,271--21\,284, 2020.

\bibitem{DCA}
L.~Wu \emph{et~al.}, ``Deep covariance alignment for domain adaptive remote sensing image segmentation,'' \emph{TGRS}, vol.~60, pp. 1--11, 2022.

\bibitem{disco}
Y.~Gao \emph{et~al.}, ``Disco: Remedying self-supervised learning on lightweight models with distilled contrastive learning,'' in \emph{ECCV}, 2022, pp. 237--253.

\bibitem{agmm}
L.~Wu \emph{et~al.}, ``Sparsely annotated semantic segmentation with adaptive gaussian mixtures,'' in \emph{CVPR}, 2023, pp. 15\,454--15\,464.

\bibitem{simsiam}
X.~Chen and K.~He, ``Exploring simple siamese representation learning,'' in \emph{CVPR}, 2021, pp. 15\,750--15\,758.

\bibitem{vit}
A.~Dosovitskiy, L.~Beyer \emph{et~al.}, ``An image is worth 16x16 words: Transformers for image recognition at scale,'' \emph{ICLR}, 2020.

\bibitem{chen2024towards}
R.~J. Chen \emph{et~al.}, ``Towards a general-purpose foundation model for computational pathology,'' \emph{Nature Medicine}, vol.~30, no.~3, pp. 850--862, 2024.

\bibitem{vorontsov2024foundation}
E.~Vorontsov \emph{et~al.}, ``A foundation model for clinical-grade computational pathology and rare cancers detection,'' \emph{Nature Medicine}, pp. 1--12, 2024.

\bibitem{vaidya2024demographic}
A.~Vaidya \emph{et~al.}, ``Demographic bias in misdiagnosis by computational pathology models,'' \emph{Nature Medicine}, vol.~30, no.~4, pp. 1174--1190, 2024.

\bibitem{nnunet}
F.~Isensee \emph{et~al.}, ``nnu-net: a self-configuring method for deep learning-based biomedical image segmentation,'' \emph{Nature Methods}, vol.~18, no.~2, pp. 203--211, 2021.

\bibitem{GLMAE}
J.-X. Zhuang \emph{et~al.}, ``Advancing volumetric medical image segmentation via global-local masked autoencoder,'' \emph{arXiv preprint arXiv:2306.08913}, 2023.

\bibitem{swinmm}
Y.~Wang \emph{et~al.}, ``Swinmm: masked multi-view with swin transformers for 3d medical image segmentation,'' in \emph{MICCAI}, 2023.

\bibitem{Mis-fm}
G.~Wang \emph{et~al.}, ``Mis-fm: 3d medical image segmentation using foundation models pretrained on a large-scale unannotated dataset,'' \emph{arXiv preprint arXiv:2306.16925}, 2023.

\bibitem{he2024foundation}
Y.~He \emph{et~al.}, ``Foundation model for advancing healthcare: Challenges, opportunities, and future directions,'' \emph{arXiv preprint arXiv:2404.03264}, 2024.

\bibitem{he2024meddr}
S.~He \emph{et~al.}, ``Meddr: Diagnosis-guided bootstrapping for large-scale medical vision-language learning,'' \emph{arXiv preprint arXiv:2404.15127}, 2024.

\bibitem{Modelgen}
Z.~Zhou \emph{et~al.}, ``Models genesis,'' \emph{MedIA}, vol.~67, p. 101840, 2021.

\bibitem{TransVW}
F.~Haghighi \emph{et~al.}, ``Transferable visual words: Exploiting the semantics of anatomical patterns for self-supervised learning,'' \emph{TMI}, vol.~40, no.~10, pp. 2857--2868, 2021.

\bibitem{wang2017chestx}
X.~Wang \emph{et~al.}, ``Chestx-ray8: Hospital-scale chest x-ray database and benchmarks on weakly-supervised classification and localization of common thorax diseases,'' in \emph{CVPR}, 2017, pp. 2097--2106.

\bibitem{chexpert}
J.~Irvin \emph{et~al.}, ``Chexpert: A large chest radiograph dataset with uncertainty labels and expert comparison,'' in \emph{AAAI}, vol.~33, no.~01, 2019, pp. 590--597.

\bibitem{meduniseg}
Y.~Ye, ``Meduniseg: 2d and 3d medical image segmentation via a prompt-driven universal model,'' \emph{arXiv preprint arXiv:2410.05905}, 2024.

\bibitem{CT}
T.~M. Buzug, ``Computed tomography,'' in \emph{Springer handbook of medical technology}, 2011, pp. 311--342.

\bibitem{withers2021x}
P.~J. Withers \emph{et~al.}, ``X-ray computed tomography,'' \emph{Nature Reviews Methods Primers}, vol.~1, no.~1, p.~18, 2021.

\bibitem{freetumor}
L.~Wu \emph{et~al.}, ``Freetumor: Advance tumor segmentation via large-scale tumor synthesis,'' 2024.

\bibitem{embracing}
Y.-C. Chou, ``Embracing massive medical data,'' in \emph{MICCAI}.\hskip 1em plus 0.5em minus 0.4em\relax Springer, 2024, pp. 24--35.

\bibitem{eva}
Y.~Fang \emph{et~al.}, ``Eva: Exploring the limits of masked visual representation learning at scale,'' in \emph{CVPR}, 2023, pp. 19\,358--19\,369.

\bibitem{stunet}
Z.~Huang \emph{et~al.}, ``Stu-net: Scalable and transferable medical image segmentation models empowered by large-scale supervised pre-training,'' 2023.

\bibitem{intra}
X.~He \emph{et~al.}, ``Intra-and inter-slice contrastive learning for point supervised oct fluid segmentation,'' \emph{TIP}, vol.~31, pp. 1870--1881, 2022.

\bibitem{rubik2}
X.~Tao \emph{et~al.}, ``Revisiting rubik’s cube: self-supervised learning with volume-wise transformation for 3d medical image segmentation,'' in \emph{MICCAI}, 2020, pp. 238--248.

\bibitem{dira}
F.~Haghighi \emph{et~al.}, ``Dira: Discriminative, restorative, and adversarial learning for self-supervised medical image analysis,'' in \emph{CVPR}, 2022, pp. 20\,824--20\,834.

\bibitem{dodnet}
J.~Zhang \emph{et~al.}, ``Dodnet: Learning to segment multi-organ and tumors from multiple partially labeled datasets,'' in \emph{CVPR}, 2021, pp. 1195--1204.

\bibitem{clipdriven}
J.~Liu \emph{et~al.}, ``Clip-driven universal model for organ segmentation and tumor detection,'' in \emph{ICCV}, 2023, pp. 21\,152--21\,164.

\bibitem{shu2023omni}
Y.~o. Shu, ``Omni-training: bridging pre-training and meta-training for few-shot learning,'' \emph{TPAMI}, 2023.

\bibitem{DBFNet}
L.~Wu \emph{et~al.}, ``Deep bilateral filtering network for point-supervised semantic segmentation in remote sensing images,'' \emph{TIP}, vol.~31, pp. 7419--7434, 2022.

\bibitem{tan2023positive}
X.~Tan \emph{et~al.}, ``Positive-negative receptive field reasoning for omni-supervised 3d segmentation,'' \emph{TPAMI}, 2023.

\bibitem{wu2024modeling}
L.~Wu \emph{et~al.}, ``Modeling the label distributions for weakly-supervised semantic segmentation,'' \emph{arXiv preprint arXiv:2403.13225}, 2024.

\bibitem{yang2022st++}
L.~Yang \emph{et~al.}, ``St++: Make self-training work better for semi-supervised semantic segmentation,'' in \emph{CVPR}, 2022, pp. 4268--4277.

\bibitem{CISC_R}
L.~Wu \emph{et~al.}, ``Querying labeled for unlabeled: Cross-image semantic consistency guided semi-supervised semantic segmentation,'' \emph{TPAMI}, vol.~45, no.~7, pp. 8827--8844, Jul. 2023.

\bibitem{yang2023revisiting}
L.~Yang \emph{et~al.}, ``Revisiting weak-to-strong consistency in semi-supervised semantic segmentation,'' in \emph{CVPR}, 2023, pp. 7236--7246.

\bibitem{liu2023multi}
Q.~Liu \emph{et~al.}, ``A multi-level label-aware semi-supervised framework for remote sensing scene classification,'' \emph{TGRS}, 2023.

\bibitem{infonce}
A.~v.~d. Oord \emph{et~al.}, ``Representation learning with contrastive predictive coding,'' \emph{arXiv preprint arXiv:1807.03748}, 2018.

\bibitem{supervised_CL}
P.~Khosla \emph{et~al.}, ``Supervised contrastive learning,'' \emph{NeurIPS}, vol.~33, pp. 18\,661--18\,673, 2020.

\bibitem{dropout}
N.~Srivastava \emph{et~al.}, ``Dropout: a simple way to prevent neural networks from overfitting,'' \emph{JMLR}, vol.~15, no.~1, pp. 1929--1958, 2014.

\bibitem{btcv}
B.~Landman \emph{et~al.}, ``Miccai multi-atlas labeling beyond the cranial vault--workshop and challenge,'' in \emph{MICCAI workshop}, vol.~5, 2015, p.~12.

\bibitem{tcia}
K.~Clark \emph{et~al.}, ``The cancer imaging archive (tcia): maintaining and operating a public information repository,'' \emph{Jour. of Dig. Imag.}, vol.~26, pp. 1045--1057, 2013.

\bibitem{luna}
A.~Setio \emph{et~al.}, ``Validation, comparison, and combination of algorithms for automatic detection of pulmonary nodules in computed tomography images: the luna16 challenge,'' \emph{MedIA}, vol.~42, pp. 1--13, 2017.

\bibitem{FLARE22}
J.~Ma \emph{et~al.}, ``Fast and low-gpu-memory abdomen ct organ segmentation: the flare challenge,'' \emph{MedIA}, vol.~82, p. 102616, 2022.

\bibitem{HNSCC}
\BIBentryALTinterwordspacing
A.~Grossberg \emph{et~al.}, ``Md anderson cancer center head and neck quantitative imaging working group,'' \emph{The Cancer Imaging Archive}, 2020. [Online]. Available: \url{https://doi.org/10.7937/k9/tcia.2020.a8sh-7363}
\BIBentrySTDinterwordspacing

\bibitem{stoic}
M.-P. Revel \emph{et~al.}, ``Study of thoracic ct in covid-19: the stoic project,'' \emph{Radiology}, vol. 301, no.~1, pp. E361--E370, 2021.

\bibitem{LIDC}
S.~G. Armato~III \emph{et~al.}, ``The lung image database consortium (lidc) and image database resource initiative (idri): a completed reference database of lung nodules on ct scans,'' \emph{Medical physics}, vol.~38, no.~2, pp. 915--931, 2011.

\bibitem{Total}
J.~Wasserthal \emph{et~al.}, ``Totalsegmentator: Robust segmentation of 104 anatomic structures in ct images,'' \emph{Radiology: Artificial Intelligence}, vol.~5, no.~5, 2023.

\bibitem{MSD}
M.~Antonelli \emph{et~al.}, ``The medical segmentation decathlon,'' \emph{Nature Commun.}, vol.~13, no.~1, p. 4128, 2022.

\bibitem{kits}
N.~Heller \emph{et~al.}, ``The kits21 challenge: Automatic segmentation of kidneys, renal tumors, and renal cysts in corticomedullary-phase ct,'' 2023.

\bibitem{chaos}
A.~E. Kavur \emph{et~al.}, ``Chaos challenge-combined (ct-mr) healthy abdominal organ segmentation,'' \emph{MedIA}, vol.~69, p. 101950, 2021.

\bibitem{panc_ct}
H.~Roth \emph{et~al.}, ``Data from pancreas-ct,'' \emph{The Cancer Imaging Archive}, 2016.

\bibitem{Difftumor}
Q.~Chen \emph{et~al.}, ``Towards generalizable tumor synthesis,'' in \emph{CVPR}, 2024.

\bibitem{word}
X.~Luo \emph{et~al.}, ``{WORD}: A large scale dataset, benchmark and clinical applicable study for abdominal organ segmentation from ct image,'' \emph{MedIA}, vol.~82, p. 102642, 2022.

\bibitem{amos}
Y.~Ji \emph{et~al.}, ``Amos: A large-scale abdominal multi-organ benchmark for versatile medical image segmentation,'' \emph{NeurIPS}, vol.~35, pp. 36\,722--36\,732, 2022.

\bibitem{DeepLesion}
\BIBentryALTinterwordspacing
M.~De~Grauw \emph{et~al.}, ``The uls23 challenge public training dataset,'' Oct. 2023. [Online]. Available: \url{https://doi.org/10.5281/zenodo.10035161}
\BIBentrySTDinterwordspacing

\bibitem{PANORAMA}
\BIBentryALTinterwordspacing
N.~Alves \emph{et~al.}, ``{The PANORAMA Study Protocol: Pancreatic Cancer Diagnosis-Radiologists Meet AI},'' Feb. 2024. [Online]. Available: \url{https://doi.org/10.5281/zenodo.10599559}
\BIBentrySTDinterwordspacing

\bibitem{OPC}
\BIBentryALTinterwordspacing
M.~L. Welch \emph{et~al.}, ``Computed tomography images from large head and neck cohort,'' \emph{The Cancer Imaging Archive}, 2023. [Online]. Available: \url{https://doi.org/10.7937/J47W-NM11}
\BIBentrySTDinterwordspacing

\bibitem{HeadNeckPET}
\BIBentryALTinterwordspacing
M.~Vallières \emph{et~al.}, ``Data from head-neck-pet-ct,'' \emph{The Cancer Imaging Archive}, 2017. [Online]. Available: \url{https://doi.org/10.7937/K9/TCIA.2017.8oje5q00}
\BIBentrySTDinterwordspacing

\bibitem{QIN-HEADNECK}
\BIBentryALTinterwordspacing
R.~R. Beichel \emph{et~al.}, ``Data from qin-headneck,'' \emph{The Cancer Imaging Archive}, 2015. [Online]. Available: \url{https://doi.org/10.7937/K9/TCIA.2015.K0F5CGLI}
\BIBentrySTDinterwordspacing

\bibitem{TCGA-HNSC}
\BIBentryALTinterwordspacing
M.~L. Zuley \emph{et~al.}, ``The cancer genome atlas head-neck squamous cell carcinoma collection (tcga-hnsc),'' \emph{The Cancer Imaging Archive}, 2016. [Online]. Available: \url{https://doi.org/10.7937/K9/TCIA.2016.LXKQ47MS}
\BIBentrySTDinterwordspacing

\bibitem{CT-COLONOGRAPHY}
\BIBentryALTinterwordspacing
S.~K \emph{et~al.}, ``Data from ct-colonography,'' \emph{The Cancer Imaging Archive}, 2015. [Online]. Available: \url{https://doi.org/10.7937/K9/TCIA.2015.NWTESAY1}
\BIBentrySTDinterwordspacing

\bibitem{MELA}
\BIBentryALTinterwordspacing
S.~Song \emph{et~al.}, ``{MELA Dataset: A Benchmark for Mediastinal Lesion Analysis},'' 2022. [Online]. Available: \url{https://doi.org/10.5281/zenodo.6575197}
\BIBentrySTDinterwordspacing

\bibitem{StonyBrookChestCT}
\BIBentryALTinterwordspacing
J.~Saltz \emph{et~al.}, ``Stony brook university covid-19 positive cases,'' \emph{The Cancer Imaging Archive}, 2021. [Online]. Available: \url{https://doi.org/10.7937/TCIA.BBAG-2923}
\BIBentrySTDinterwordspacing

\bibitem{CT-CLIP}
I.~E. Hamamci \emph{et~al.}, ``A foundation model utilizing chest ct volumes and radiology reports for supervised-level zero-shot detection of abnormalities,'' \emph{arXiv preprint arXiv:2403.17834}, 2024.

\bibitem{NLST}
\BIBentryALTinterwordspacing
N.~L. S. T.~R. Team, ``Data from the national lung screening trial (nlst),'' \emph{The Cancer Imaging Archive}, 2013. [Online]. Available: \url{https://doi.org/10.7937/TCIA.HMQ8-J677}
\BIBentrySTDinterwordspacing

\bibitem{wang2024towards}
H.~Wang and X.~Li, ``Towards generic semi-supervised framework for volumetric medical image segmentation,'' \emph{NeurIPS}, vol.~36, 2024.

\bibitem{mmwhs}
X.~Zhuang, ``Multivariate mixture model for myocardial segmentation combining multi-source images,'' \emph{TPAMI}, vol.~41, no.~12, pp. 2933--2946, 2018.

\bibitem{avt}
L.~Radl \emph{et~al.}, ``Avt: Multicenter aortic vessel tree cta dataset collection with ground truth segmentation masks,'' \emph{Data in brief}, vol.~40, p. 107801, 2022.

\bibitem{SLIVER07}
\BIBentryALTinterwordspacing
B.~van Ginneken, ``Sliver07 [data set],'' Mar. 2019. [Online]. Available: \url{https://doi.org/10.5281/zenodo.2597908}
\BIBentrySTDinterwordspacing

\bibitem{3D-IRCADb}
\BIBentryALTinterwordspacing
L.~Soler \emph{et~al.}, ``3d image reconstruction for comparison of algorithm database,'' \emph{Data}, 2010. [Online]. Available: \url{https://www.ircad.fr/research/data-sets/liver-segmentation-3d-ircadb-01/}
\BIBentrySTDinterwordspacing

\bibitem{Kipa}
Y.~He \emph{et~al.}, ``Meta grayscale adaptive network for 3d integrated renal structures segmentation,'' \emph{MedIA}, vol.~71, p. 102055, 2021.

\bibitem{segthor}
Z.~Lambert, C.~Petitjean, B.~Dubray, and S.~Kuan, ``Segthor: Segmentation of thoracic organs at risk in ct images,'' in \emph{Inter. Conf. Image Process. Theory Tools. Appli.}, 2020, pp. 1--6.

\bibitem{BHSD}
B.~Wu \emph{et~al.}, ``Bhsd: A 3d multi-class brain hemorrhage segmentation dataset,'' in \emph{Inter. Workshop Machine Learn. Medical Imag.}, 2023, pp. 147--156.

\bibitem{StructSeg}
\BIBentryALTinterwordspacing
J.~Shi, ``Structseg2019 gtv segmentation,'' \emph{IEEE Dataport}, 2023. [Online]. Available: \url{https://dx.doi.org/10.21227/h75x-gt46}
\BIBentrySTDinterwordspacing

\bibitem{verse}
A.~Sekuboyina \emph{et~al.}, ``Verse: A vertebrae labelling and segmentation benchmark for multi-detector ct images,'' \emph{MedIA}, vol.~73, p. 102166, 2021.

\bibitem{covid}
H.~R. Roth \emph{et~al.}, ``Rapid artificial intelligence solutions in a pandemic—the covid-19-20 lung ct lesion segmentation challenge,'' \emph{MedIA}, vol.~82, p. 102605, 2022.

\bibitem{FUMPE}
M.~Masoudi \emph{et~al.}, ``A new dataset of computed-tomography angiography images for computer-aided detection of pulmonary embolism,'' \emph{Scientific data}, vol.~5, no.~1, pp. 1--9, 2018.

\bibitem{Parse22}
G.~Luo \emph{et~al.}, ``Efficient automatic segmentation for multi-level pulmonary arteries: The parse challenge,'' \emph{arXiv preprint arXiv:2304.03708}, 2023.

\bibitem{AIIB23}
Y.~Nan \emph{et~al.}, ``Fuzzy attention neural network to tackle discontinuity in airway segmentation,'' \emph{TNNLS}, 2023.

\bibitem{CC-CCII}
K.~Zhang \emph{et~al.}, ``Clinically applicable ai system for accurate diagnosis, quantitative measurements, and prognosis of covid-19 pneumonia using computed tomography,'' \emph{Cell}, vol. 181, no.~6, pp. 1423--1433, 2020.

\bibitem{AutoPET}
S.~Gatidis \emph{et~al.}, ``A whole-body fdg-pet/ct dataset with manually annotated tumor lesions,'' \emph{Scientific Data}, vol.~9, no.~1, p. 601, 2022.

\bibitem{ACDC}
O.~Bernard \emph{et~al.}, ``Deep learning techniques for automatic mri cardiac multi-structures segmentation and diagnosis: is the problem solved?'' \emph{TMI}, vol.~37, no.~11, pp. 2514--2525, 2018.

\bibitem{ATLAS-MRI}
F.~Quinton \emph{et~al.}, ``A tumour and liver automatic segmentation (atlas) dataset on contrast-enhanced magnetic resonance imaging for hepatocellular carcinoma,'' \emph{Data}, vol.~8, no.~5, p.~79, 2023.

\bibitem{brats}
U.~Baid \emph{et~al.}, ``The rsna-asnr-miccai brats 2021 benchmark on brain tumor segmentation and radiogenomic classification,'' \emph{arXiv preprint arXiv:2107.02314}, 2021.

\bibitem{cyclemorph}
B.~Kim \emph{et~al.}, ``Cyclemorph: cycle consistent unsupervised deformable image registration,'' \emph{MedIA}, vol.~71, p. 102036, 2021.

\bibitem{marcus2007open}
D.~S. Marcus \emph{et~al.}, ``Open access series of imaging studies (oasis): cross-sectional mri data in young, middle aged, nondemented, and demented older adults,'' \emph{Journal of Cognit. Neur.}, vol.~19, no.~9, pp. 1498--1507, 2007.

\bibitem{CTRG}
Y.~Tang \emph{et~al.}, ``Work like a doctor: Unifying scan localizer and dynamic generator for automated computed tomography report generation,'' \emph{Expert Sys. Appli.}, vol. 237, p. 121442, 2024.

\bibitem{swinunetr}
A.~Hatamizadeh \emph{et~al.}, ``Swin unetr: Swin transformers for semantic segmentation of brain tumors in mri images,'' in \emph{MICCAI Workshop}, 2021, pp. 272--284.

\bibitem{Monai}
M.~J. Cardoso \emph{et~al.}, ``Monai: An open-source framework for deep learning in healthcare,'' \emph{arXiv preprint arXiv:2211.02701}, 2022.

\bibitem{unetr}
A.~Hatamizadeh \emph{et~al.}, ``Unetr: Transformers for 3d medical image segmentation,'' in \emph{WACV}, 2022, pp. 574--584.

\bibitem{UNET}
O.~Ronneberger, P.~Fischer, and T.~Brox, ``U-net: Convolutional networks for biomedical image segmentation,'' in \emph{MICCAI}, 2015, pp. 234--241.

\bibitem{jigsaw}
P.~Chen \emph{et~al.}, ``Jigsaw clustering for unsupervised visual representation learning,'' in \emph{CVPR}, 2021, pp. 11\,526--11\,535.

\bibitem{Transmorph}
J.~Chen \emph{et~al.}, ``Transmorph: Transformer for unsupervised medical image registration,'' \emph{MedIA}, vol.~82, p. 102615, 2022.

\bibitem{voxelmorph}
G.~Balakrishnan \emph{et~al.}, ``Voxelmorph: a learning framework for deformable medical image registration,'' \emph{TMI}, vol.~38, no.~8, pp. 1788--1800, 2019.

\bibitem{siebert2021fast}
H.~Siebert \emph{et~al.}, ``Fast 3d registration with accurate optimisation and little learning for learn2reg 2021,'' in \emph{MICCAI}, 2021, pp. 174--179.

\bibitem{mok2021conditional}
T.~C. Mok and A.~C. Chung, ``Conditional deformable image registration with convolutional neural network,'' in \emph{MICCAI}, 2021, pp. 35--45.

\bibitem{bleu}
K.~Papineni \emph{et~al.}, ``Bleu: a method for automatic evaluation of machine translation,'' in \emph{ACL}, 2002, pp. 311--318.

\bibitem{Mesh-Memor}
M.~Cornia \emph{et~al.}, ``Meshed-memory transformer for image captioning,'' in \emph{CVPR}, 2020, pp. 10\,578--10\,587.

\bibitem{rstnet}
X.~Zhang \emph{et~al.}, ``Rstnet: Captioning with adaptive attention on visual and non-visual words,'' in \emph{CVPR}, June 2021, pp. 15\,465--15\,474.

\bibitem{GSKET}
S.~Yang \emph{et~al.}, ``Knowledge matters: Chest radiology report generation with general and specific knowledge,'' \emph{MedIA}, vol.~80, p. 102510, 2022.

\bibitem{m2kt}
------, ``Radiology report generation with a learned knowledge base and multi-modal alignment,'' \emph{MedIA}, vol.~86, p. 102798, 2023.

\bibitem{freemask}
L.~Yang \emph{et~al.}, ``Freemask: Synthetic images with dense annotations make stronger segmentation models,'' \emph{NeurIPS}, vol.~36, 2024.

\end{thebibliography}
\bibliographystyle{IEEEtran}
\vspace{-.6in}

\end{document}